\documentclass[conference,compsoc]{IEEEtran}

\ifCLASSOPTIONcompsoc
  \usepackage[nocompress]{cite}
\else
  \usepackage{cite}
\fi

\usepackage{booktabs}

\ifCLASSINFOpdf
  \usepackage[pdftex]{graphicx}
\else
\fi
\usepackage{amsmath}

\usepackage{algorithmic}
\usepackage{algorithm}
\usepackage{multirow}
\usepackage{textcase}
\usepackage[caption=false,font=footnotesize,labelfont=sf,textfont=sf]{subfig}
\begin{document}
%
% paper title
% Titles are generally capitalized except for words such as a, an, and, as,
% at, but, by, for, in, nor, of, on, or, the, to and up, which are usually
% not capitalized unless they are the first or last word of the title.
% Linebreaks \\ can be used within to get better formatting as desired.
% Do not put math or special symbols in the title.
\title{Addressing the Shape-Radiance Ambiguity in View-Dependent Radiance Fields}

% author names and affiliations
% use a multiple column layout for up to three different
% affiliations
\author{\IEEEauthorblockN{Sverker Rasmuson}
\IEEEauthorblockA{Chalmers University of Technology\\
Gothenburg, Sweden \\
Email: sverker.rasmuson@chalmers.se}
\and
\IEEEauthorblockN{Erik Sintorn}
\IEEEauthorblockA{Chalmers University of Technology\\
Gothenburg, Sweden\\
Email: erik.sintorn@chalmers.se}
\and
\IEEEauthorblockN{Ulf Assarsson}
\IEEEauthorblockA{Chalmers University of Technology\\
Gothenburg, Sweden\\
Email: uffe@chalmers.se}}

% make the title area
\maketitle

\begin{figure*}[h!]
\subfloat{\includegraphics[width=0.24\linewidth]{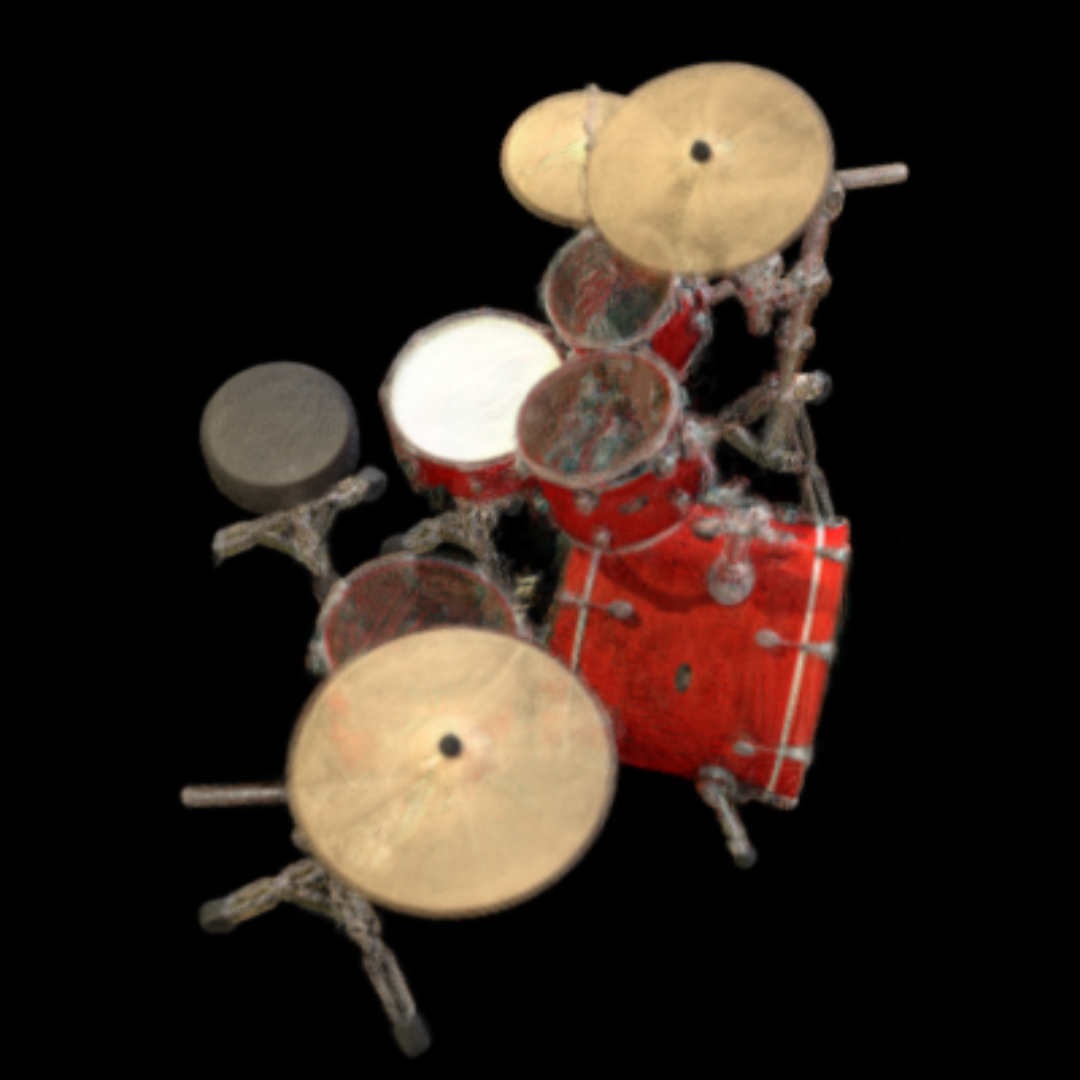}}
\subfloat{\includegraphics[width=0.24\linewidth]{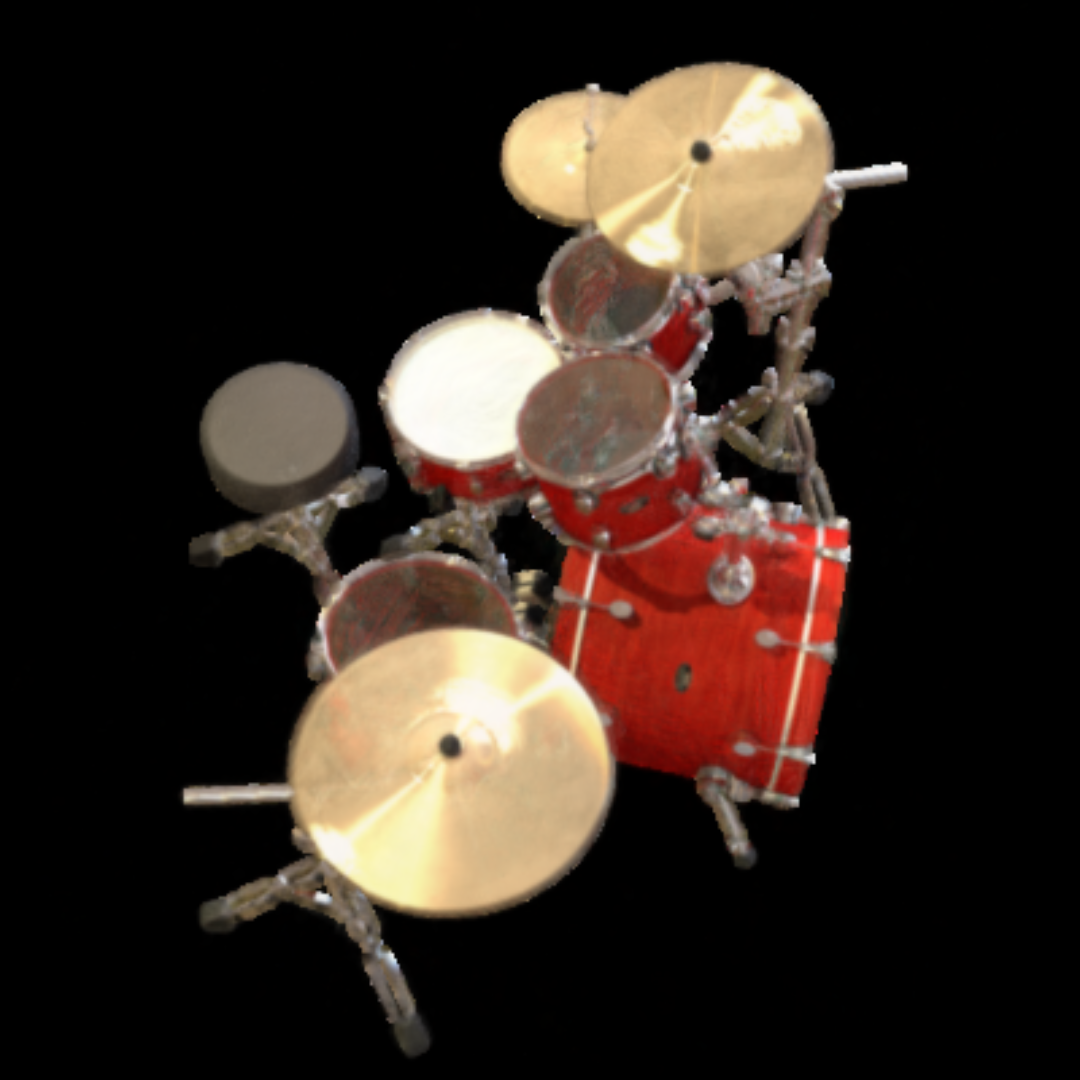}}
\subfloat{\includegraphics[width=0.24\linewidth]{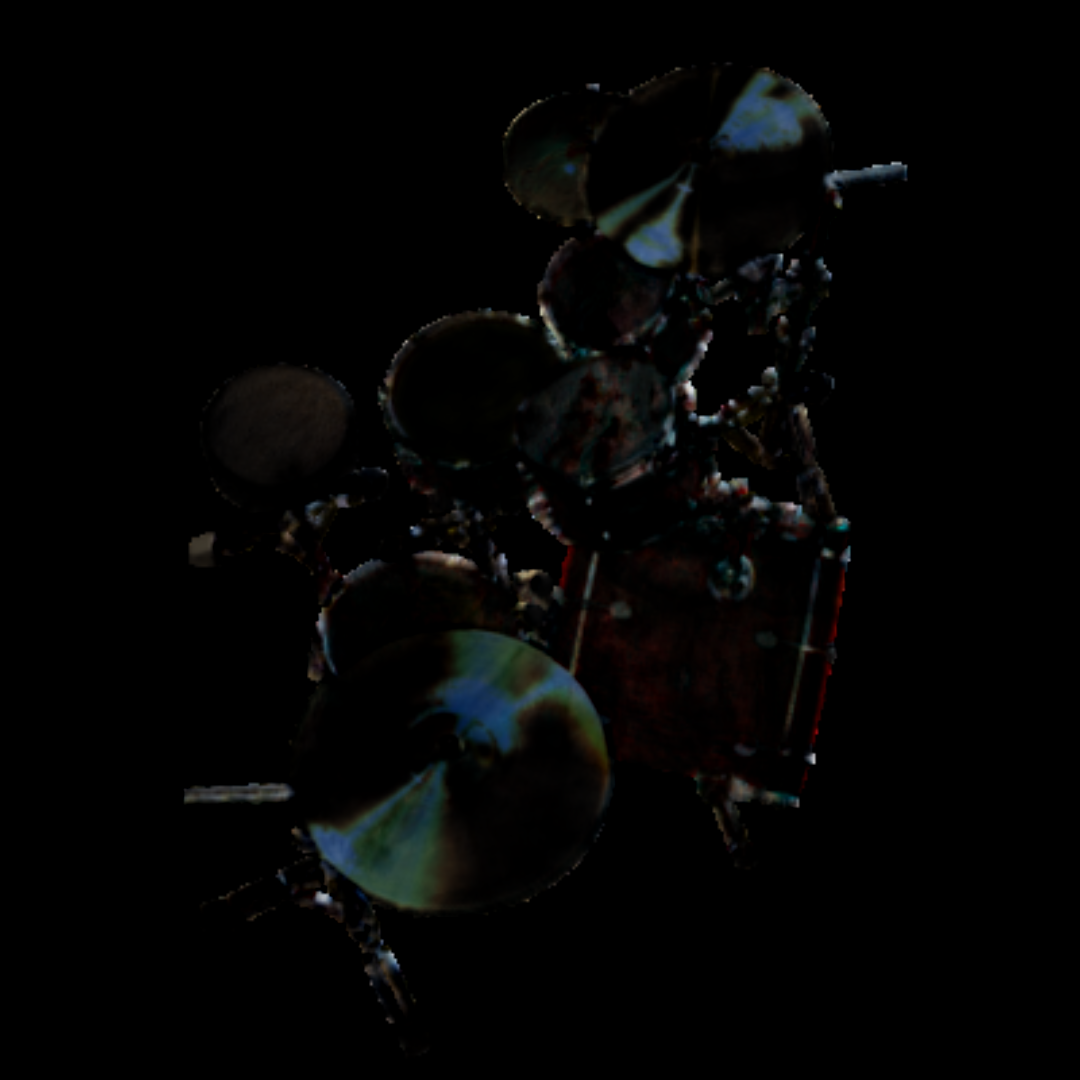}}
\subfloat{\includegraphics[width=0.24\linewidth]{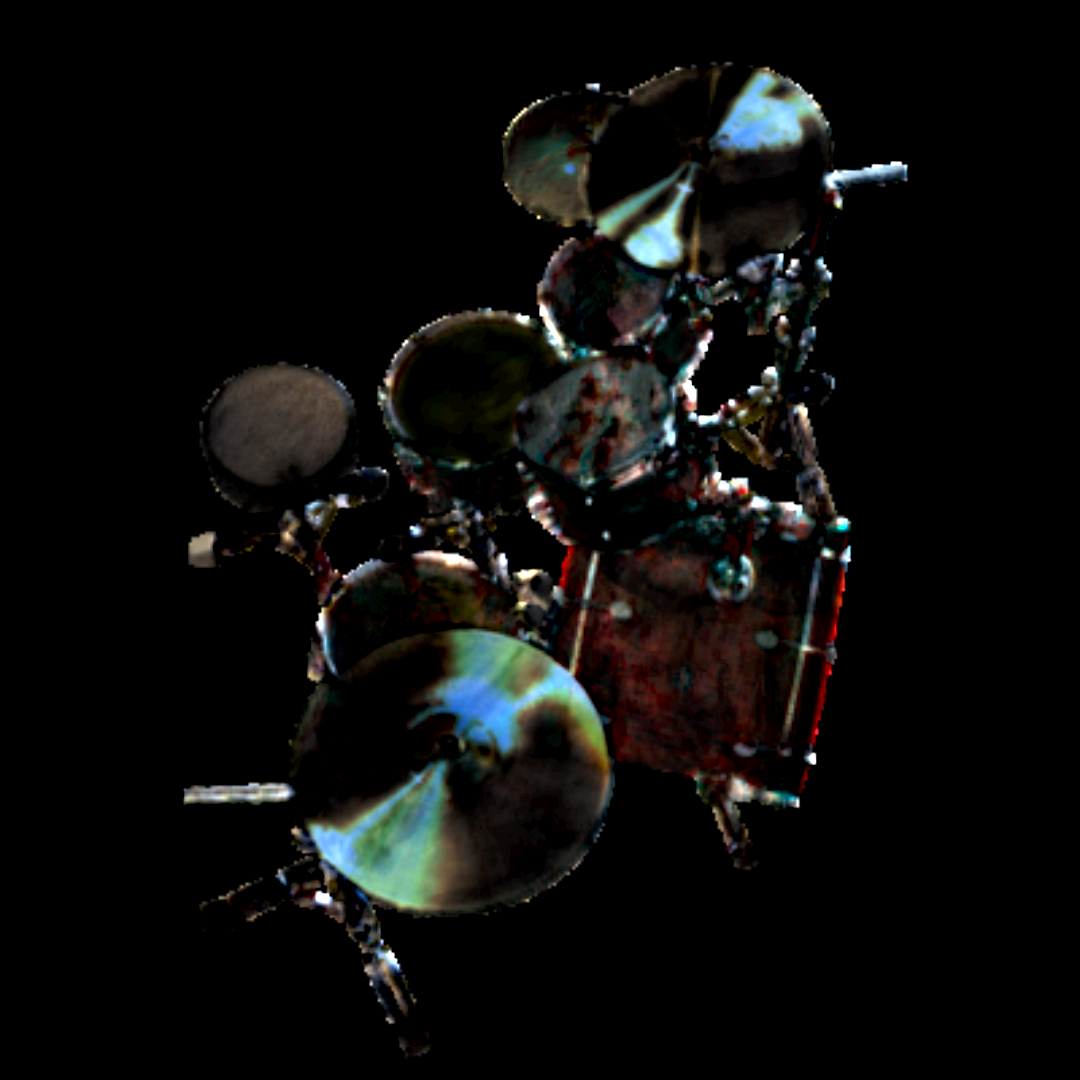}}
\caption{Results  from  our  algorithm on the drums data set\cite{Mildenhall2020}.  From  left  to  right:  Lambertian color, 
Lambertian + view-dependent color, view-dependent color only, view-dependent color brightened for easier inspection.}
\label{fig:teaser}
\end{figure*}

% As a general rule, do not put math, special symbols or citations
% in the abstract
\begin{abstract}
We present a method for handling view-dependent information in radiance fields to help with convergence and quality of 3D reconstruction. Radiance fields 
with view-dependence suffers from
the so called shape-radiance ambiguity, which can lead to incorrect geometry given a high angular resolution of view-dependent colors. 
We propose the addition of a difference plane in front of each camera, with the purpose of 
separating view-dependent and Lambertian components during training. We also propose an additional step where we train, but do not store, a low-resolution 
view-dependent function that helps to isolate the surface if such a separation is proven difficult.
These additions have a small impact on performance and memory usage but enables 
reconstruction of scenes with highly specular components without any other explicit handling of view-dependence such as
Spherical Harmonics.
\end{abstract}

\IEEEpeerreviewmaketitle

\section{Introduction}
\noindent Recently, NERF~\cite{Mildenhall2020} and its successors have shown unparalleled results of novel view synthesis of near-photorealistic quality, including 
view-dependent effects. One pitfall of this approach is the shape-radiance 
ambiguity, where a case of perfect modeling of view-dependent colors leads to a scenario where every scene can be modelled by, e.g., a sphere, i.e., information about
geometry is lost~\cite{zhang2020nerf}. The model of view-dependence thus has to be good enough to be visually plausible, while not
over-fitting the data so that it cannot generalize to new views. While NERF handles most view-dependent effects admirably, there are situations
that can be problematic. Storing and computing view-dependent functions can also incur significant overhead to performance and memory.

To alleviate this problem, we propose a method that aims to separate view-dependent and Lambertian colors during reconstruction. In this 
method, the information in the
volume is considered to be purely Lambertian, while the view-dependent components of each camera is captured in a \textit{difference plane}
in front of each camera. The colors in the volume cannot be considered Lambertian while semi-transparent, but will get closer 
to this approximation when the volume elements converge to opaque or nearly opaque surfaces. The difference planes store a
single value per pixel, which contain information on how much the current
pixel deviates from the integrated Lambertian content of each ray sent through the volume (see Figure~\ref{fig:teaser}). 

The reasoning behind this approach is that if a pixel's color significantly deviates from the Lambertian colors accumulated through the 
volume, it is by definition view-dependent, and needs to be handled separately.

In some cases it can be hard to separate the colors in purely Lambertian and view-dependent components. We therefore
propose the addition of a constraint where a small view-dependent function is computed per voxel. The parameters of this function
are not stored, but are only used to increase or decrease the density per voxel to make convergence to the correct surface easier.

A general downside of methods based on neural rendering are the long
reconstruction times; usually up to a day or more for a typical scene. Recent papers, such as Plenoxels\cite{yu2021plenoxels}, DirectVoxGo\cite{sun2021direct} and 
PERF\cite{rasmuson2021perf}, have shown significant improvements of reconstruction times by solving the
optimization problem directly instead of invoking neural networks. To accommodate for view-dependent effects, Plenoxels store Spherical Harmonics 
for each 
voxel cell. With 9 extra coefficients per voxel this incur a significant memory cost. DirectVoxGo instead store a grid of features which are evaluated on a shallow MLP. This also requires extra storage and
adds a penalty to training and rendering time. 

Our method adds low overhead to the PERF-method, and still enables us to handle scenes with highly view-dependent components. 
This scene can then be used as is or augmented with view-dependent information. The existing geometry should help mitigate the shape-radiance
ambiguity in such an endeavor.

\section{Related work}
\noindent Here, we will cover the most relevant previous work, divided into separate categories. 

\paragraph{\textbf{Multi-view Stereo}} One classical approach to 3D reconstruction is Multi-View Stereo (MVS), where a set of images are taken of a scene from multiple 
viewpoints. The images are then paired in sequence to compute depth information about the scene via triangulation. MVS has been 
well researched over the years, for an overview see Hartley and Zisserman~\cite{Zisserman2004} and Seitz
et al.~\cite{seitz2006comparison}. Newer research have to coupled this method with neural networks in Deep Multi-view 
Stereo~\cite{yao2018}\cite{yao2019}.

\paragraph{\textbf{Structure from Motion}}
Another classical approach in Computer Vision is to reconstruct scenes with Structure from Motion (SfM), where image features are used
to associate images with another, typically with SIFT 
features~\cite{Lowe:2004:DIF:993451.996342}\cite{snavely2006photo}\cite{ozyecsil2017}. 3d positions corresponding to image features are 
jointly optimized with camera parameters to reconstruct the scene. Typically such a reconstruction is sparse, but it can be used as the 
basis for a dense reconstruction using e.g. PatchMatch Stereo~\cite{Bleyer2011}\cite{schonberger2016structure}.

\paragraph{\textbf{Novel View Synthesis}}
Novel View Synthesis is related to 3D reconstruction but focuses on generating novel viewpoints instead of reconstructing the underlying 
geometry. Well researched areas include image-based rendering and lightfields~\cite{shum2000}. 

Multi-Plane Images (MPIs) is a recent approach where neural networks are used to train semi-transparent images that lie in different 
planes between the cameras and the scene. These can then be rendered from novel view points by blending the contributions of each such 
plane intersecting with the view-rays from a virtual camera. ~\cite{Mildenhall2019}\cite{Zhou2018}\cite{Broxton2020}.

\paragraph{\textbf{Neural rendering}}
The interest in neural rendering, where neural networks are coupled with a volume representation of the scene, has exploded in recent 
years. Neural rendering allows for great reconstruction quality
of general scenes, especially when it comes to novel view synthesis, as shown in Neural Volumes~\cite{Lombardi2019} and 
NERF~\cite{Mildenhall2020}. NERF uses positional encoding for its inputs to the neural network, which give better results for
high frequency scene information~\cite{Tancik2020a}.

Plenty of research has followed this work, for example improving upon the rendering times and view 
synthesis quality~\cite{martinbrualla2020}\cite{liu2020}\cite{yu2021}\cite{barron2021mipnerf}. An overview of this field can be found
in the STAR-paper by Tewari et al.~\cite{Tewari2020NeuralSTAR}.

Nerf++\cite{zhang2020nerf} shows that a key component to the ability for NERF to generalize so well to novel views is the structure
of the Multi Layer Perceptron (MLP), where the view direction is inserted late in the network. This regularizes the colors to vary 
smoothly with view direction, and  there are less parameters to describe the overall view-dependent color, avoiding over-fitting.

A downside of neural rendering methods are the long times it takes to train the MLP, which can amount to days. Early attempts to speed
up training use pretraining or combine it with known methods such as external MVS reconstructions 
~\cite{Tancik2020}\cite{Chen2021}\cite{liu2021neuray}.

Recently, Plenoxels\cite{yu2021plenoxels}, DirectVoxGo\cite{sun2021direct} and PERF\cite{rasmuson2021perf} have proposed to speed up
the reconstruction by direct optimization instead of using neural networks. This improves the reconstruction times
by many orders of magnitude compared to the original NERF implementation. View-dependent effects are modeled  with Spherical 
Harmonics in Plenoxels, which incur a significant memory cost. DirectVoxGo use a shallow MLP to account for these effects in a similar
way as SNeRG~\cite{hedman2021snerg}, which induces a penalty to training and rendering times.

Our work is built as an addition to the non-linear least squares framework in PERF~\cite{rasmuson2021perf}, which has no explicit handling of view-dependent 
effects. We aim to keep the performance benefits of this framework while still ameliorating the problem of the shape-radiance ambiguity.

\section{Shape-Radiance Ambiguity}
\begin{figure}[!h]
\centering
\includegraphics[width=0.2\textwidth]{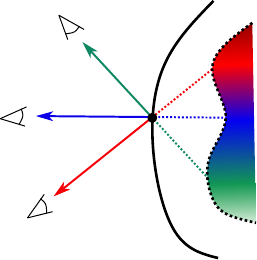}
\label{fig:ambiguity}
\caption{Illustration of the shape-radiance ambiguity. A single point on a false surface (solid) can satisfy the incoming radiance to the cameras
through view-dependent colors, even though these rays correspond to three distinct points on the real surface (dotted). }
\end{figure}

\noindent Granted enough angular resolution of view-dependent colors, almost any shape can satisfy the incoming radiance to a given camera. This is easy to realize
by studying Figure~\ref{fig:ambiguity}, where a single (erroneous) point satisfies the incoming radiance for three different views. In reality, they actually
correspond to three distinct points on the surface.

\begin{figure}[!t]
\subfloat{\includegraphics[width=0.45\linewidth]{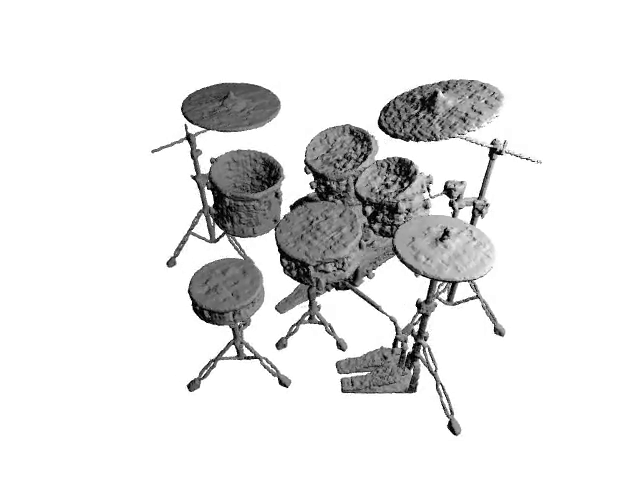}}
\subfloat{\includegraphics[width=0.45\linewidth]{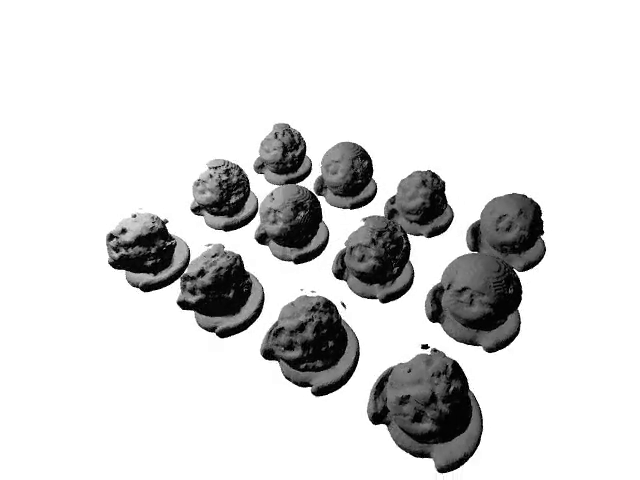}}
\caption{Geometry from reconstruction with NERF, converted to triangles via marching cubes. Artifacts from the shape-radiance ambiguity can be
seen, especially on surfaces where the color is highly view-dependent. }
\label{fig:iso}
\end{figure}

In Nerf++\cite{zhang2020nerf}, the authors suggest that the structure of the MLP in NERF serves to regularize the view-dependent colors. This in combination with the limited angular
resolution available (the size of the view-dependent MLP) works to ameliorate the problem of the shape-radiance ambiguity for novel view synthesis.
However, artifacts can still be seen if the goal instead is to recover the reconstructed geometry. In Figure~\ref{fig:iso}, 
an example of the geometry from a scene reconstructed with NERF is shown, where the radiance field is converted to triangles via marching cubes. The geometry is
especially problematic in areas that correspond to highly view-dependent surfaces.

This example shows that additional constraints are needed for NERF and other methods built on view-dependent
radiance fields to work as an end-to-end pipeline for 3D reconstruction. In this paper, we propose two low-overhead mechanisms of separating view-dependent and Lambertian color information for this purpose.

\section{Method}
\noindent As in PERF we represent the volume with a voxel grid, storing color and density per cell. The scene is surrounded with a hierarchical shell of 
environment maps to help with foreground/background segmentation.

\subsection{Volume Rendering}
\noindent The volume is rendered by, for each pixel in a virtual camera, traversing a ray through the volume and integrating the resulting
color according to
\begin{equation}
    \mathbf{H} = \sum_t T(t)(1-e^{(-\sigma(t)\delta(t))})\mathbf{c}(t)
    \label{eq:H}
\end{equation}
for step $t$, where
\begin{equation*}
    \label{eq:alpha}
   T(t) = e^{(-\sum_{t_n}^{t}\sigma(t)\delta(t))}
\end{equation*}
is the accumulated throughput, $\sigma$ is the density, $\delta$ is the distance between two adjacent steps, and $\mathbf{c}$ is the color.

\subsection{Volume Reconstruction}
\noindent The volume is constructed by invoking Equation~\ref{eq:H} for all pixels in the collection of images, and comparing
the accumulated color to the reference color at the corresponding pixel. This amounts to minimizing the objective function
$F$ with respect to all density and color values $(c_j; \sigma_j)$ such that
\begin{align}
    \min_{c_j;\sigma_j}F&, \\
    F &= 0.5 \sum_i \sum_{k = 1}^3(H_{i,k} - r_{i,k})^2
    \label{eq:F}
\end{align}
for all pixels $i$ and each color channel $k$.

\subsection{Difference planes}
To each camera image we associate a difference plane with identical resolution. For this plane we store one float value
per pixel. This value, $\alpha_{s_i}$, represents how to weight in a view-dependent part of the camera image according to:
\begin{equation}
    \hat{\mathbf{H}}_i = (1.0 - e^{(-\alpha_{s_i}\sigma_s)})(\mathbf{r}_i - \mathbf{H}_{d_i}) + \mathbf{H}_{d_i}
    \label{eq:Hi}
\end{equation}
where $r_i$ is the reference color in the corresponding pixel $i$, and

\begin{align*}
    \mathbf{H}_{d_i} &= \sum_j e^{-V_j}(1.0 - e^{(-\sigma_j\delta_j)})\mathbf{c}_j, \\
    V_j &= \sum_{k = 0}^{j-1}\sigma_k\delta_k
\end{align*}

\begin{figure}[!h]
\centering
\includegraphics[width=0.2\textwidth]{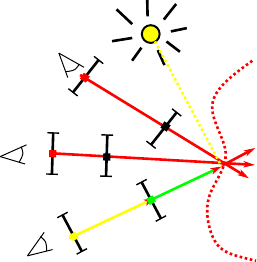}
\label{fig:planes}
\caption{Illustration of how the difference planes can capture a deviating color at a certain point, in this case
induced from a highlight seen only from a specific angle. The deviating camera sees a yellow color (from the light source) in its
reference image, compared to most other cameras which see a red diffuse color. The camera can be satisfied by adding green to its 
corresponding difference plane. The diffuse color in the volume can thus stay red while still satisfying all cameras.}
\end{figure}

 \noindent is the Lambertian part of the color accumulated from the ray traversing the volume through voxels $j$ according to Equation~\ref{eq:H}. The variable $\alpha_s$ is set to zero
initially, but can grow given that no acceptable solution can be found using only the Lambertian values in the volume, as part of the minimization of Equation~\ref{eq:F}.
The parameter $\sigma_s$ can be used as a weight to tune how readily values should be pushed into the specular planes. This value
is constant, with a value of $\sigma_s = 0.002$ for all of our experiments.

The value to be minimized, in the same manner as equation~\ref{eq:F}, is the squared differences of all accumulated values $\hat{H_i}$ and the reference color $r_i$ such that 
\begin{equation}
\hat{F} = 0.5\sum_i \sum_{k = 1}^3(\hat{H}_{i,k} - r_{i,k})^2
\label{eq:Fhat}
\end{equation}

\noindent for all pixels $i$ and each color channel $k$.

\subsection{Partial derivatives}
\noindent To allow for this minimization, partial derivatives of equation~\ref{eq:Fhat} need to be computed. For all voxel colors $c_j$ we get

\begin{equation*}
\frac{\partial F}{\partial\mathbf{c}_j} = \sum_i(\mathbf{H}_i - \mathbf{r}_i)\frac{\partial\mathbf{H_i}}{\partial\mathbf{c}_j}
\label{eq:Fpartial}
\end{equation*}
using the chain rule. The last factor can be expanded in the same manner as

\begin{equation*}
    \frac{\partial\mathbf{H_i}}{\partial\mathbf{c}_j} = e^{(-\alpha_{s_i}\sigma_s)}\frac{\partial\mathbf{H}_{d_i}}{\partial\mathbf{c}_j}.
\end{equation*}

This means that the partial derivatives $\frac{\partial\mathbf{H}_d}{\partial\mathbf{c}_j}$ can  be computed in the same manner as in PERF. The same derivation 
also holds for the density values $\alpha_j$ and the corresponding partial derivatives $\frac{\partial\mathbf{H}_d}{\partial\alpha_j}$.

The partial derivative with respect to the new variable $\alpha_{s_i}$ trivially becomes

\begin{equation*}
    \frac{\partial\mathbf{H_i}}{\partial\alpha_{s_i}} = -\sigma_s e^{(-\alpha_{s_i}\sigma_s)}(\mathbf{r}_i - \mathbf{H}_{d_i}).
\end{equation*}

\subsection{Non-highlight colors}
\noindent To be able to handle view-dependent lighting that does not behave like a highlight, i.e., does not have a clear Lambertian and view-dependent component, we have added an additional step to our method. 

NERF and similar methods can capture geometry by utilizing a low-pass filtering function to describe view dependence.
If this low-resolution function can find a good approximation of the outgoing radiance for a given point in space, it
is more likely that this is a point on a surface than an arbitrary point in the volume.

To emulate this behavior, keeping in mind our goal of performant 3D reconstruction, we define a low-frequency view dependent function for each voxel. This function is implemented as a small environment map of $8x4$ pixels. For each voxel, we populate this environment map by blending colors from all cameras, first ray-marching through the volume to account for visibility, see algorithm~\ref{algo:pop}.

\begin{algorithm}[ht]
\caption{Algorithm for populating the small view-dependent function for each voxel. Executed once for each voxel.} \label{algo:pop}
\begin{algorithmic}
\STATE $v = \text{current voxel}$
\FOR{all cameras $\mathbf{c}$}
\STATE $\mathbf{p} = \text{project\_voxel}(v, \mathbf{c})$
\STATE $T = \text{trace\_visibility}(v, \mathbf{c})$
\STATE $(\theta, \phi) = \text{map\_spherical}(v, \mathbf{c})$
\STATE $\text{env\_map}(\theta, \phi) \mathrel{+}= T\mathbf{p} $
\STATE $T_{sum} \mathrel{+}= T$
\ENDFOR \\
\FOR{$\theta\in[0, \pi], \phi\in[-\pi, \pi]$}
\STATE $\text{env\_map}(\theta, \phi) \mathrel{/}= T_{sum}$
\ENDFOR \\
\end{algorithmic}
\end{algorithm}

This function is then evaluated on how well it succeeded to approximate every given camera color with visibility, by taking
the squared error between the reference color in each camera and the color in the environment map for the corresponding direction.
For each voxel we get an error

\begin{equation}
E = \frac{1}{\sum_k T_k}\sum_k \sum_i T_k (c_i - p_i)^2 
\label{eq:cf}
\end{equation}
for visibility (to each camera $k$) $T_k$, voxel color $c_i$ and predicted color $p_i$ for all cameras $k$ and color channels $i$.

We have now obtained a cost function for each voxel in the grid, which gives us a measure of how likely it is that a given point
in space lies on a surface. 

This cost function is added as a weight $w$ to the Cauchy loss following Plenoxels \cite{yu2021plenoxels} and SnerG \cite{hedman2021snerg}:

\begin{equation}
\mathcal{L}_c = \lambda_c \sum_i log(1 + \lambda_n w \sigma_i^2)
\label{eq:cauchy}
\end{equation}
for density $\sigma$ for each voxel $i$, where $\lambda_c$ and $\lambda_n$ controls the overall loss and how much the cost function
should weigh in respectively. We use the values $\lambda_c = 0.05$ for the artifical scenes and $\lambda_c = 0.01$ for the real 
scenes, with $\lambda_n = 10$ for both. 

This equation is trivially differentiable, and can be implemented in the non-linear least squares framework of PERF by adding 
a residual per density value $\sigma$ to be minimized.

\begin{figure}[!h]
\subfloat{\includegraphics[width=\linewidth]{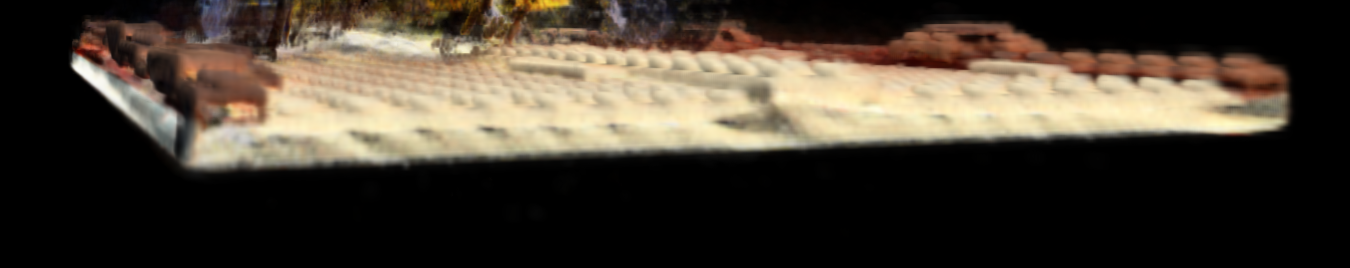}}

\subfloat{\includegraphics[width=\linewidth]{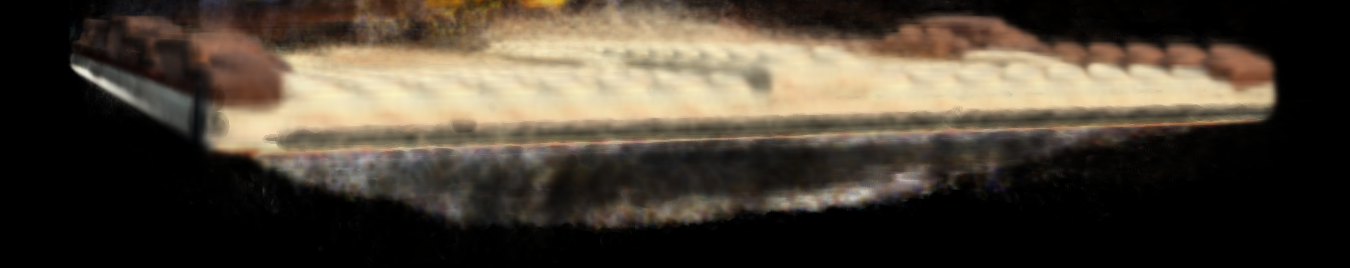}}

\label{fig:detail}
\caption{Comparison of a reconstructed detail in the lego scene when using our handling of non-highlight colors (top) compared
to a baseline (bottom).}
\end{figure}

With this implementation no expensive view-dependent function needs to be explicitly stored, while the benefits of finding correct
geometry are still obtained (see Figure~\ref{fig:detail}).

\subsection{Auxiliary terms}
\noindent To ease the computational impact of equation Algorithm~\ref{algo:pop} and Equation~\ref{eq:cf}, we implement a simple
method to skip empty space using the hierarchical formulation in PERF\cite{rasmuson2021perf}. For each level above the first, 
we store the voxel grid of the previous level. If the density value for a voxel in that grid is smaller then a threshold
$\lambda_v = 1.0$, then that voxel is skipped and treated as if being empty.

For the real scenes from the Tanks and Temples data set\cite{Knapitsch2017}, we also use a quadratic on the total visibility $T$ for each
ray following PERF\cite{rasmuson2021perf}:
\begin{equation}
   \mathcal{L}_s = \lambda_s(-4(T - 0.5)^2 + 1)
\end{equation}
with $\lambda_s = 0.1$. This term helps to mitigate smoke-like artifacts and keeping the real scenes sparse.

\section{Implementation}\label{sec:impl}
\noindent The difference planes of Equation~\ref{eq:Hi} and Cauchy loss with per-voxel weight of Equation~\ref{eq:cauchy} are added to the 
non-linear least-squares framework used in PERF. 
% As a consequence of the update formulation for the ray integration in equation~\ref{eq:Hi}, three additional residuals are added per ray.

The resolution of the added planes follow the camera resolutions through the hierarchical steps, so that it approximately matches the voxel resolution 
at any given level of the hierarchy. The values of each new hierarchical level are upsampled using linear interpolation.

The scenes are calibrated using the Structure from Motion software Colmap~\cite{schonberger2016structure}. 

Algorithm~\ref{algo:pop} and Equation~\ref{eq:cf} are run once in the beginning of each hierarchical level (except the first), and the
weights, one float per voxel, are stored in a buffer. The Cauchy loss in Eqaution~\ref{eq:cauchy} is then evaluated in each iteration 
with these weights.

\section{Results}
\begin{figure}[!t]
\subfloat{\includegraphics[width=0.24\linewidth]{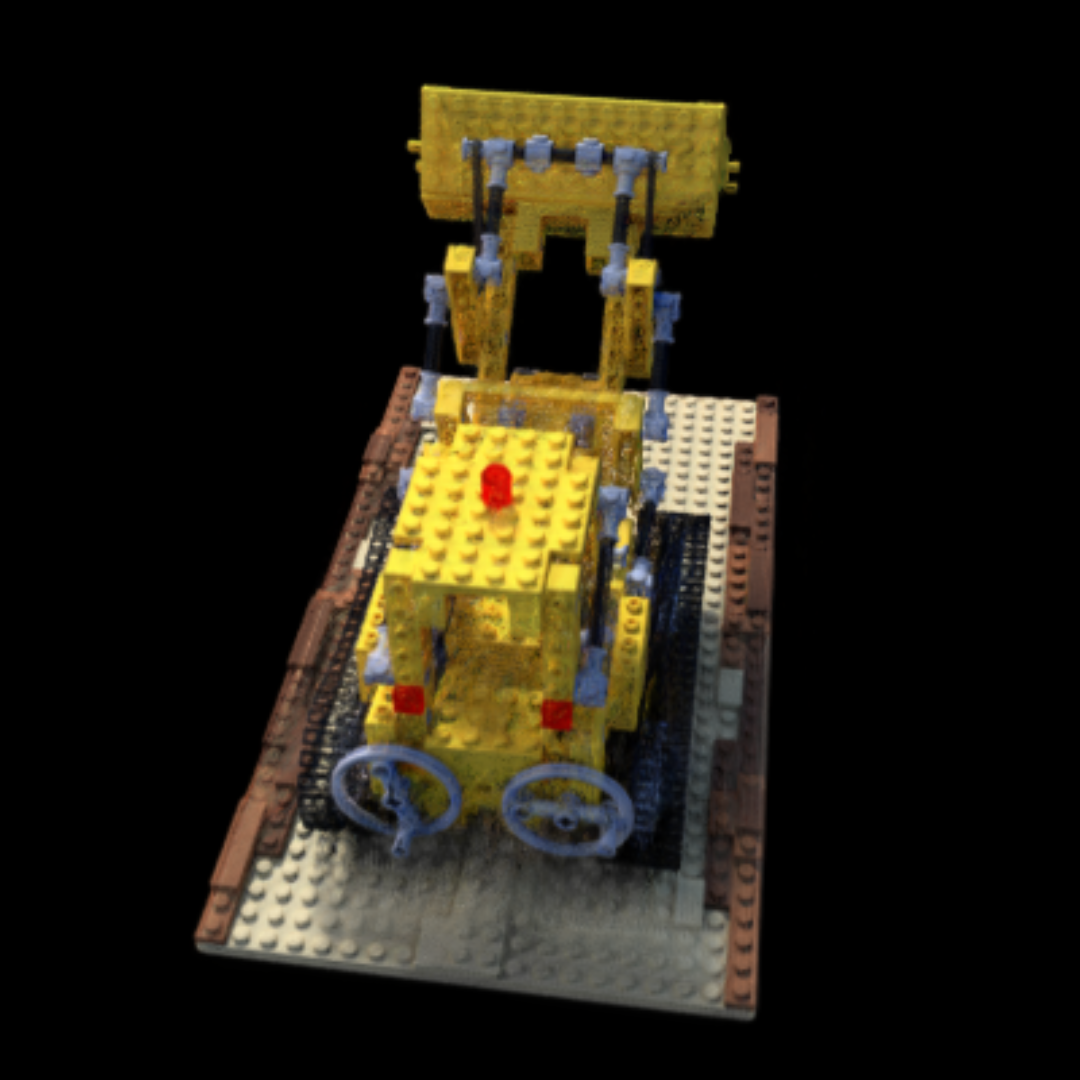}}
\subfloat{\includegraphics[width=0.24\linewidth]{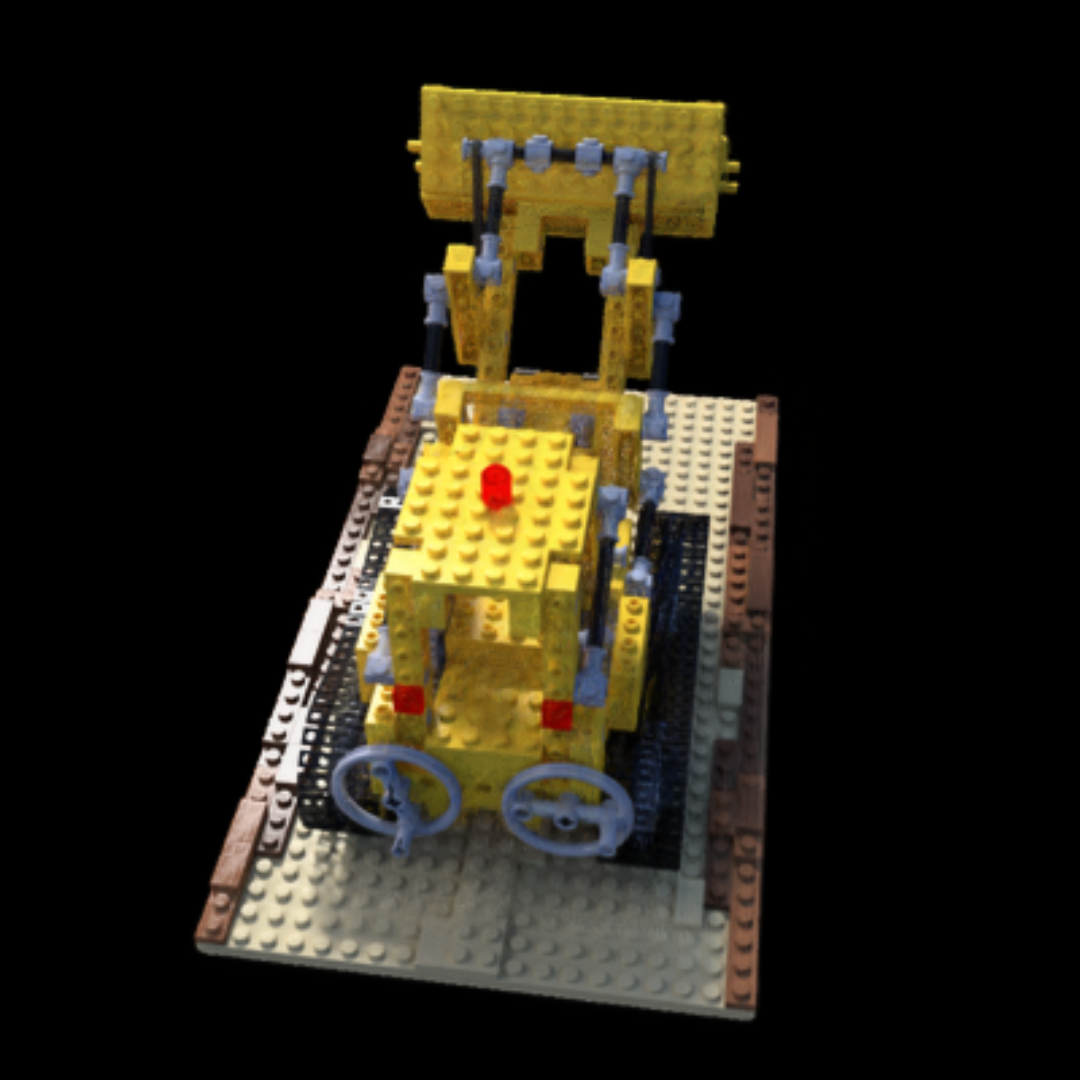}}
\subfloat{\includegraphics[width=0.24\linewidth]{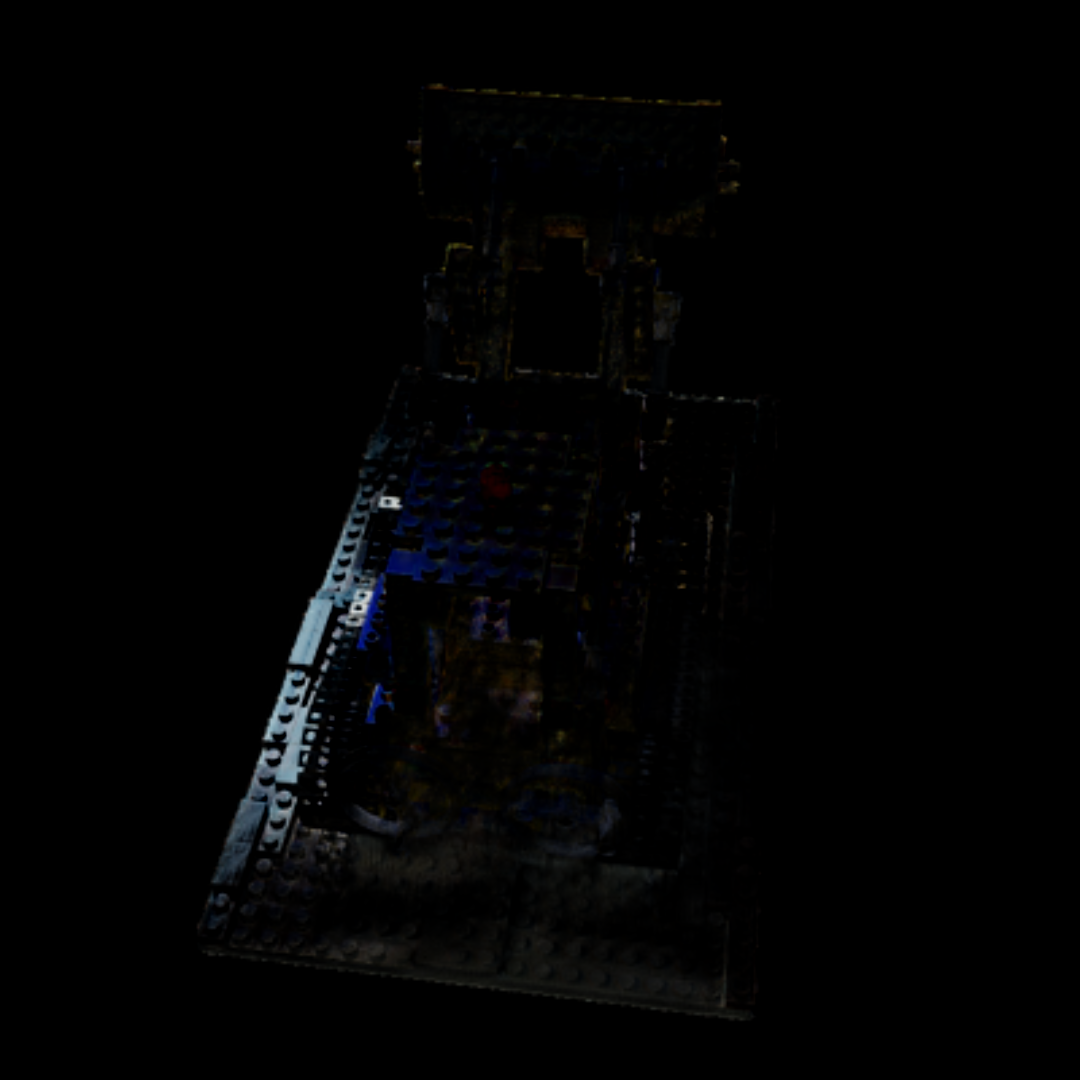}}
\subfloat{\includegraphics[width=0.24\linewidth]{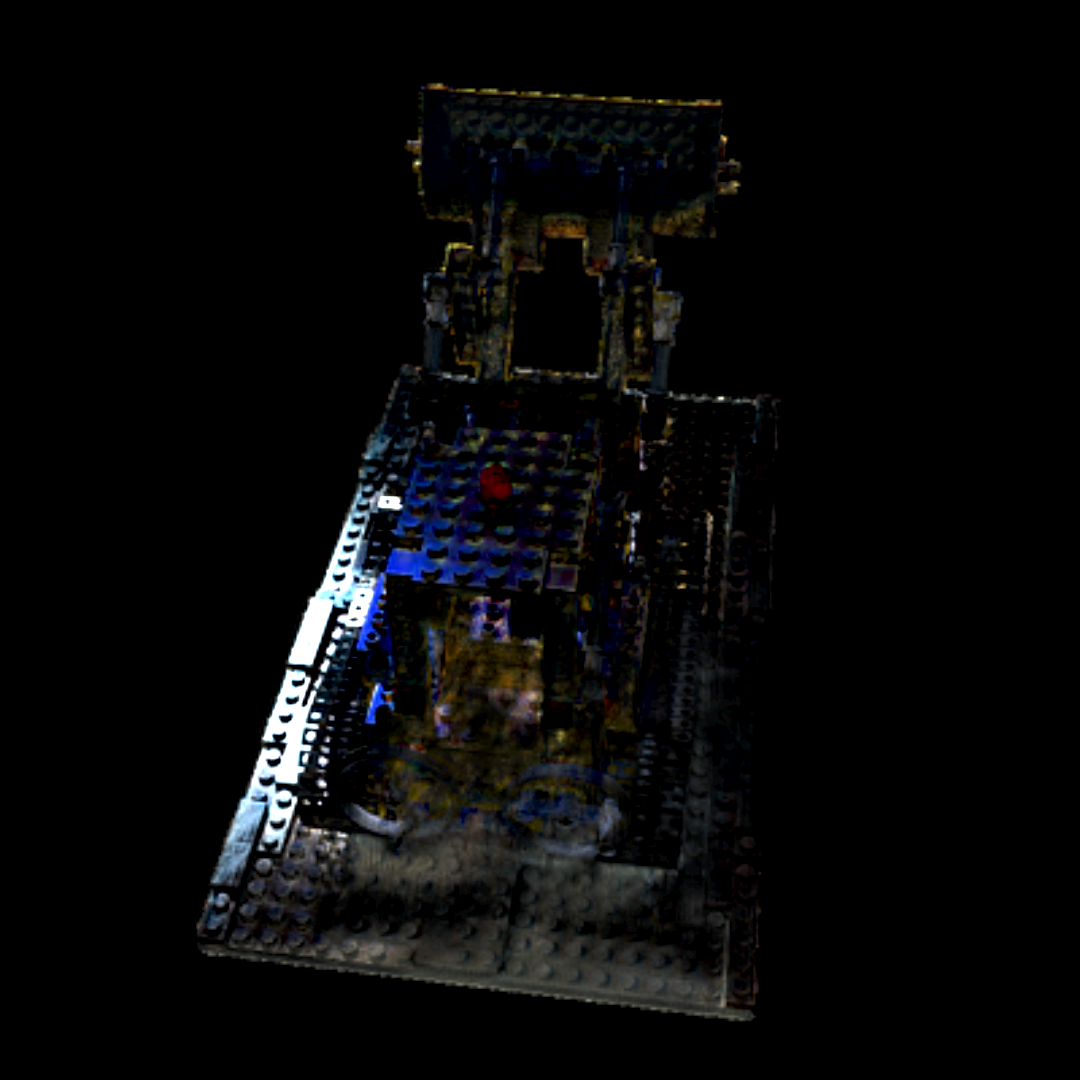}}

\subfloat{\includegraphics[width=0.24\linewidth]{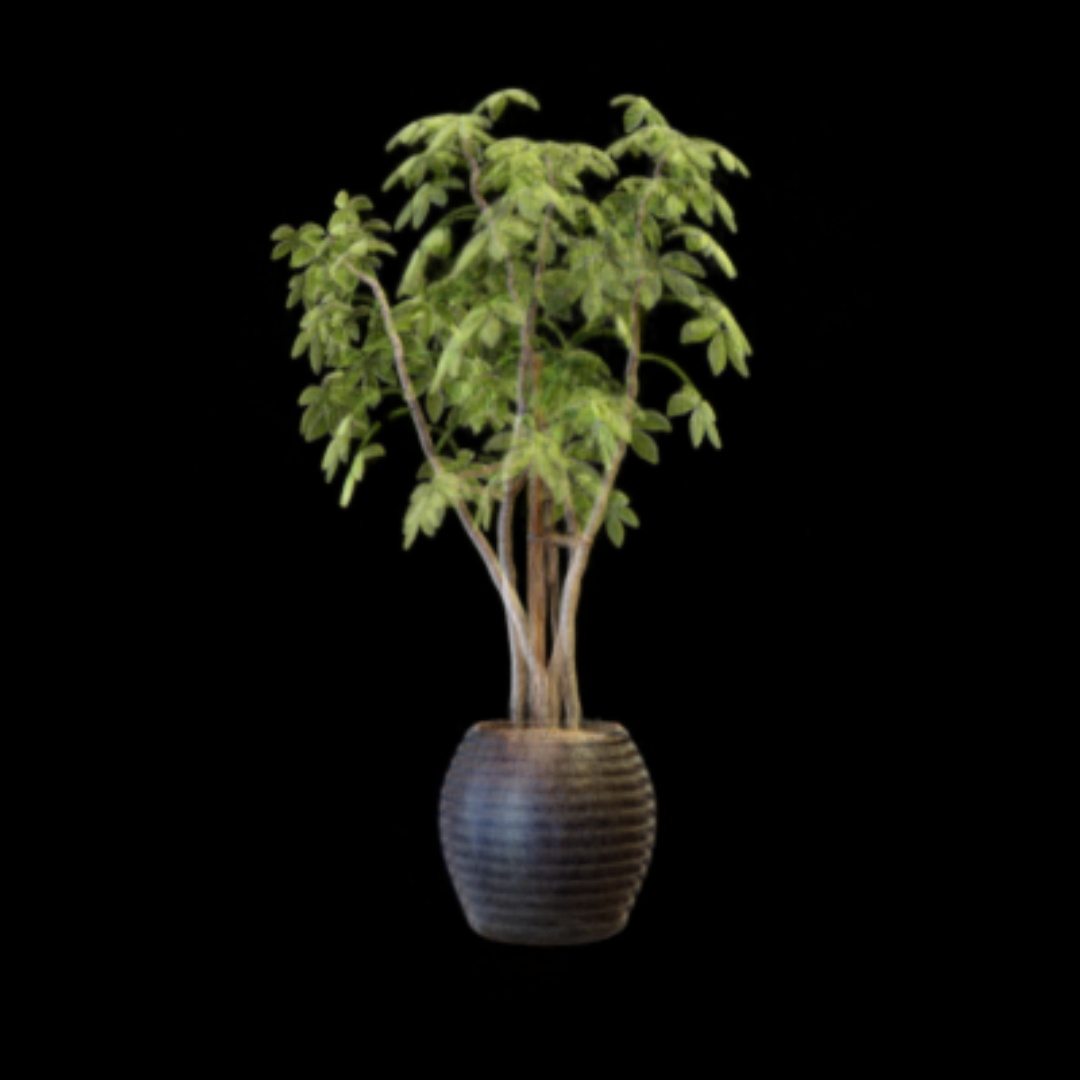}}
\subfloat{\includegraphics[width=0.24\linewidth]{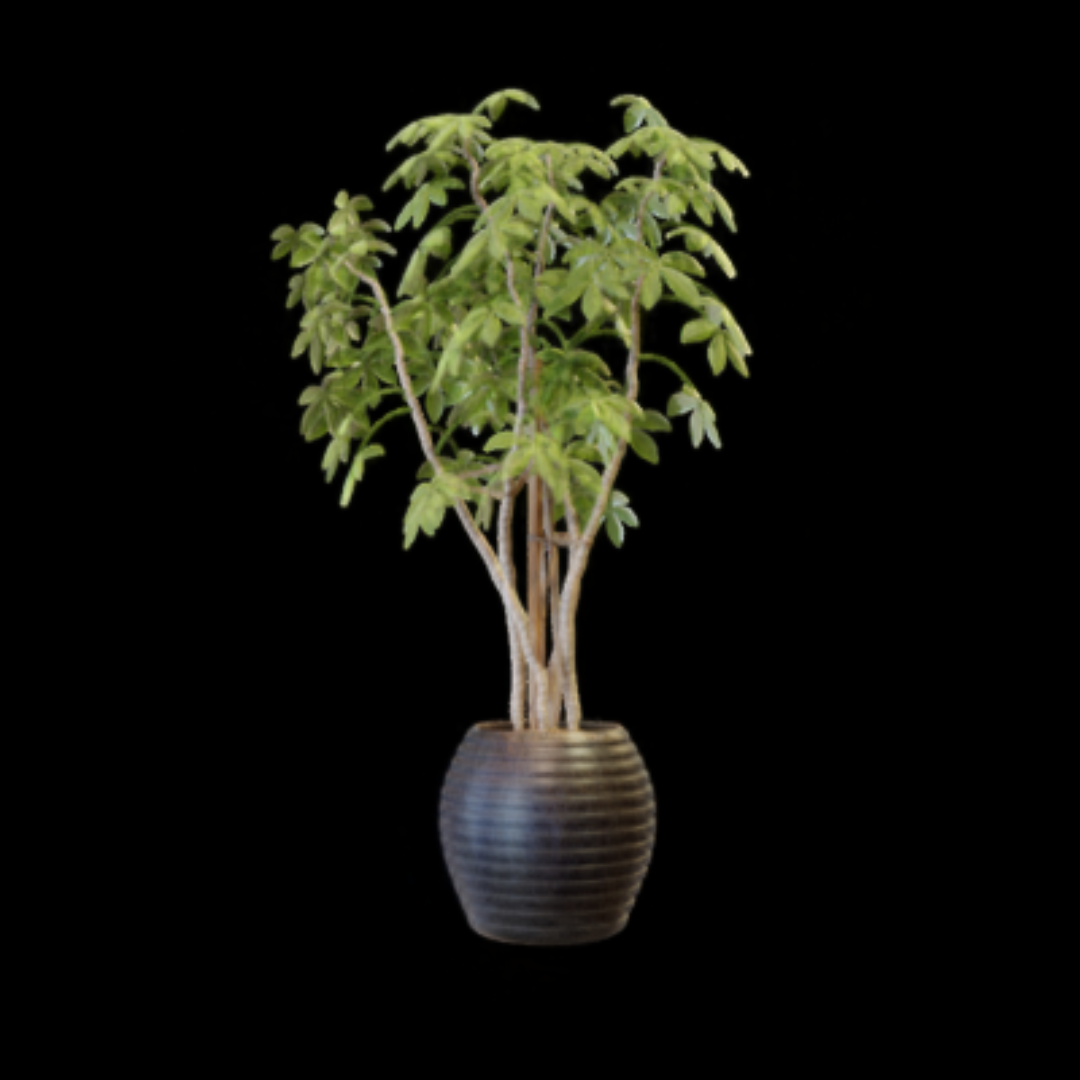}}
\subfloat{\includegraphics[width=0.24\linewidth]{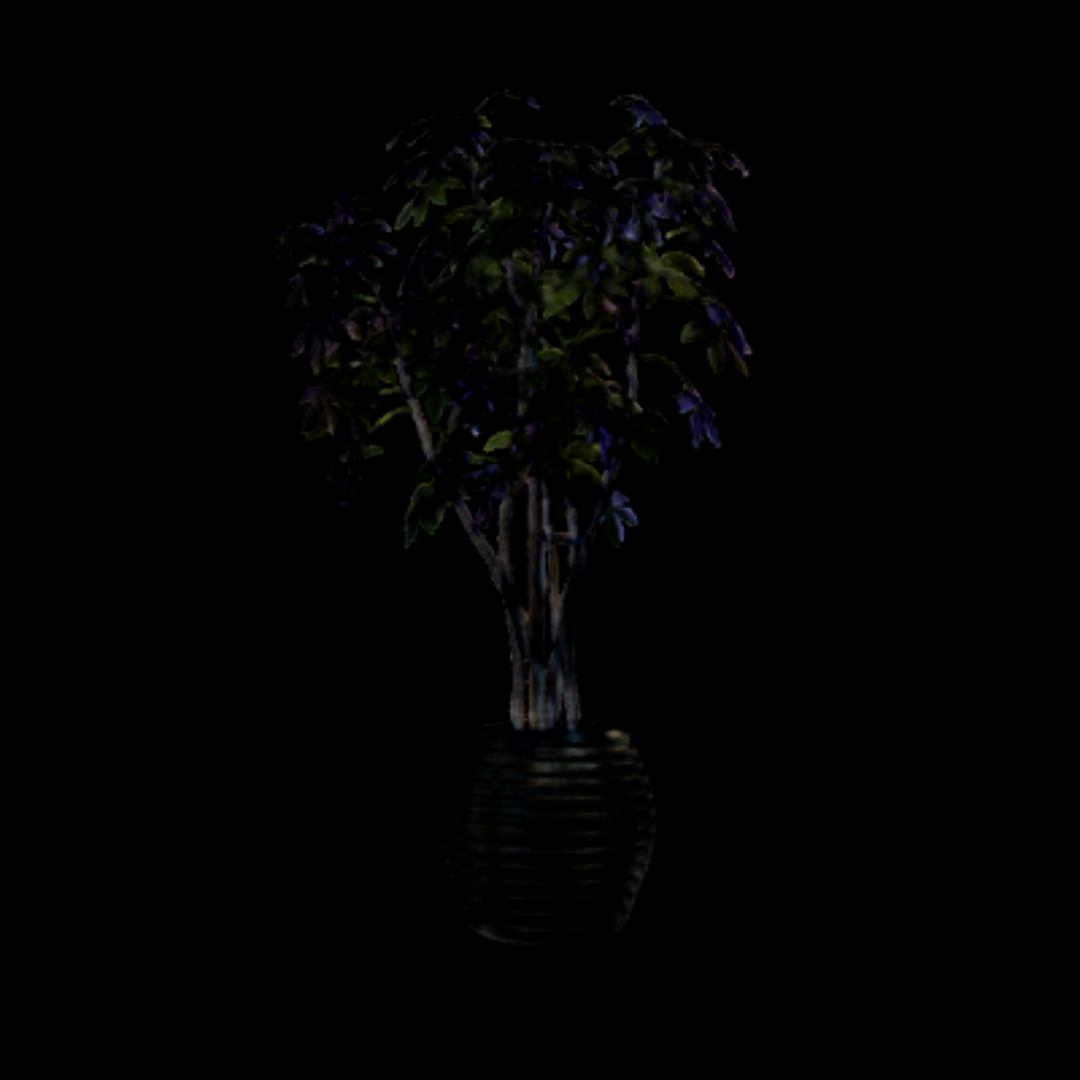}}
\subfloat{\includegraphics[width=0.24\linewidth]{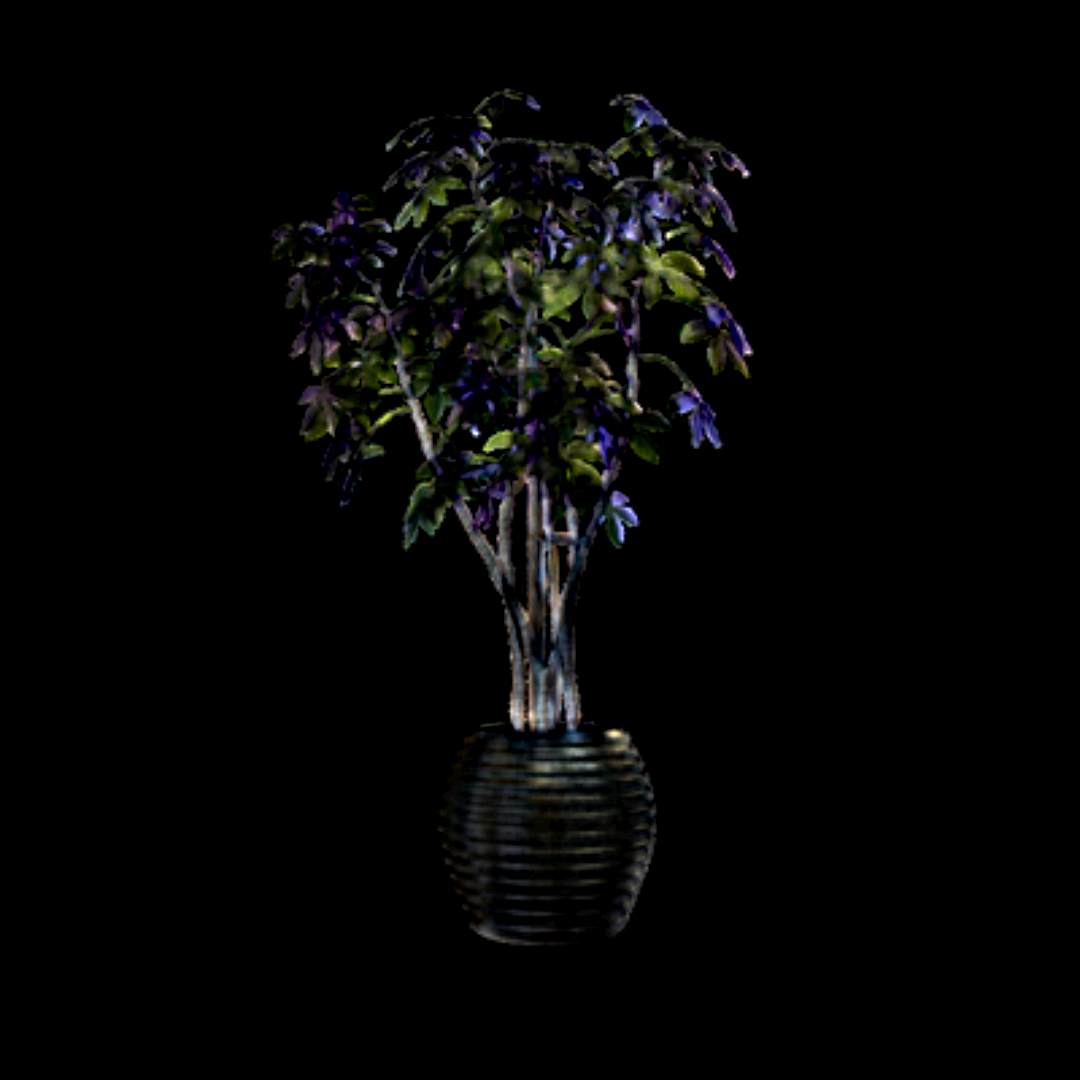}}

\subfloat{\includegraphics[width=0.24\linewidth]{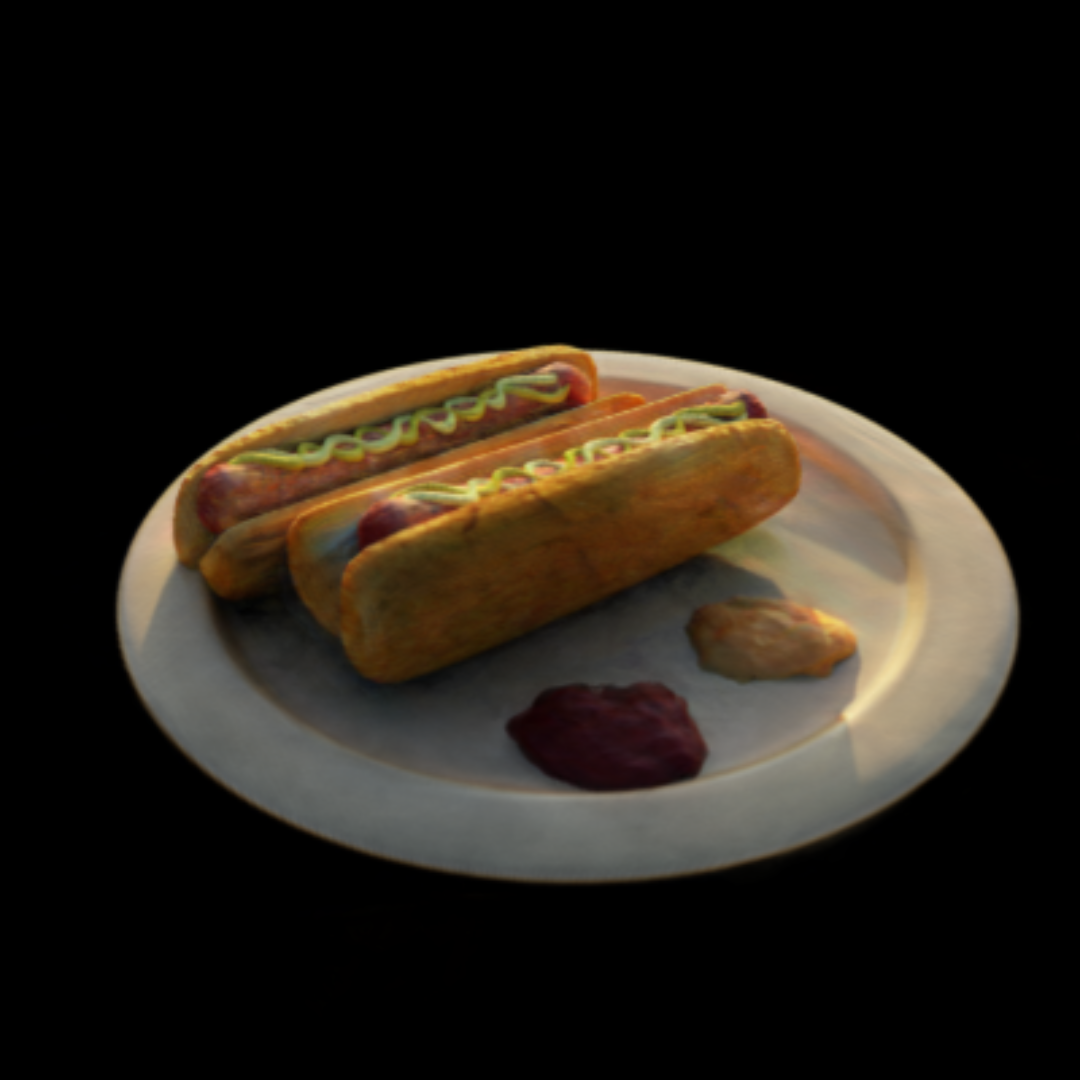}}
\subfloat{\includegraphics[width=0.24\linewidth]{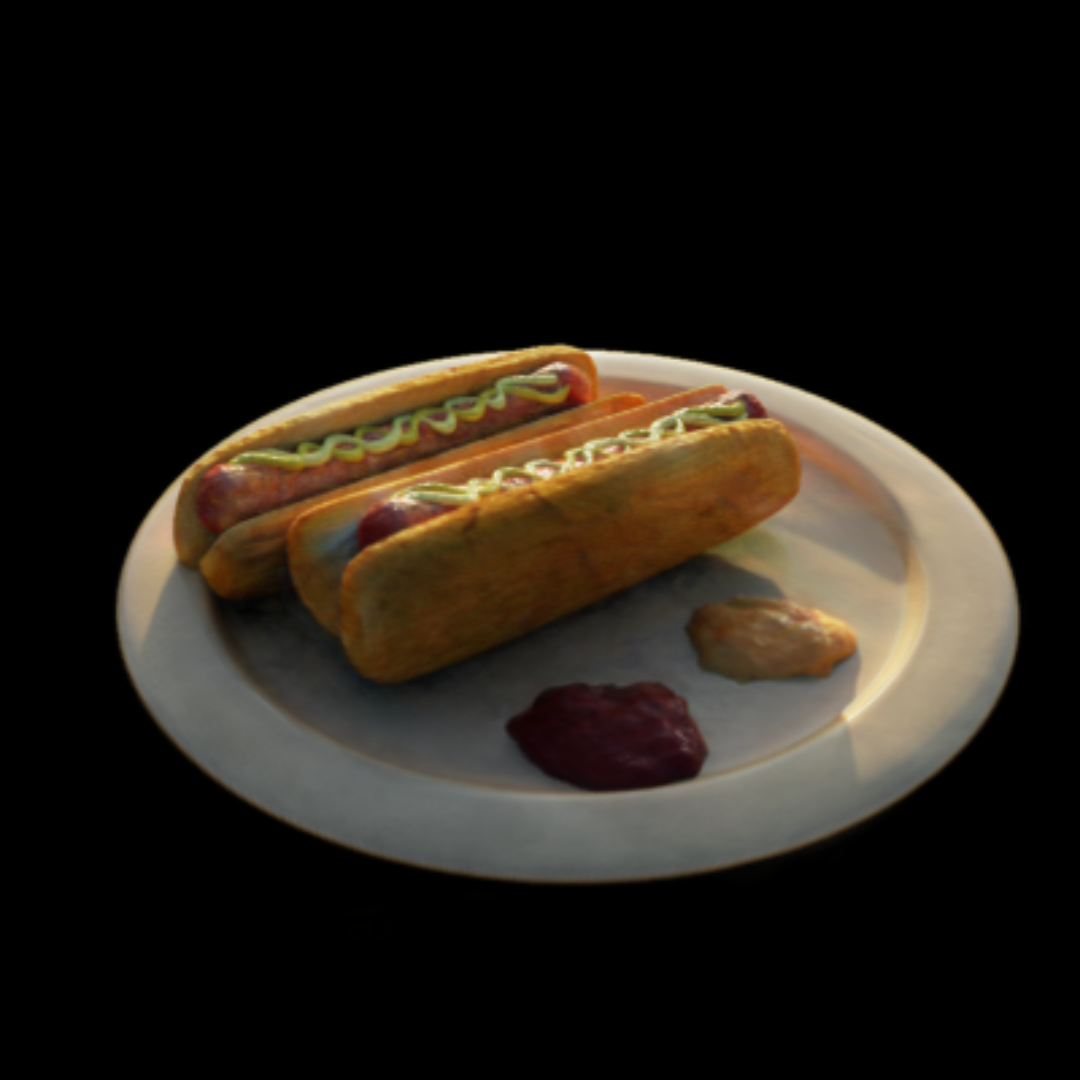}}
\subfloat{\includegraphics[width=0.24\linewidth]{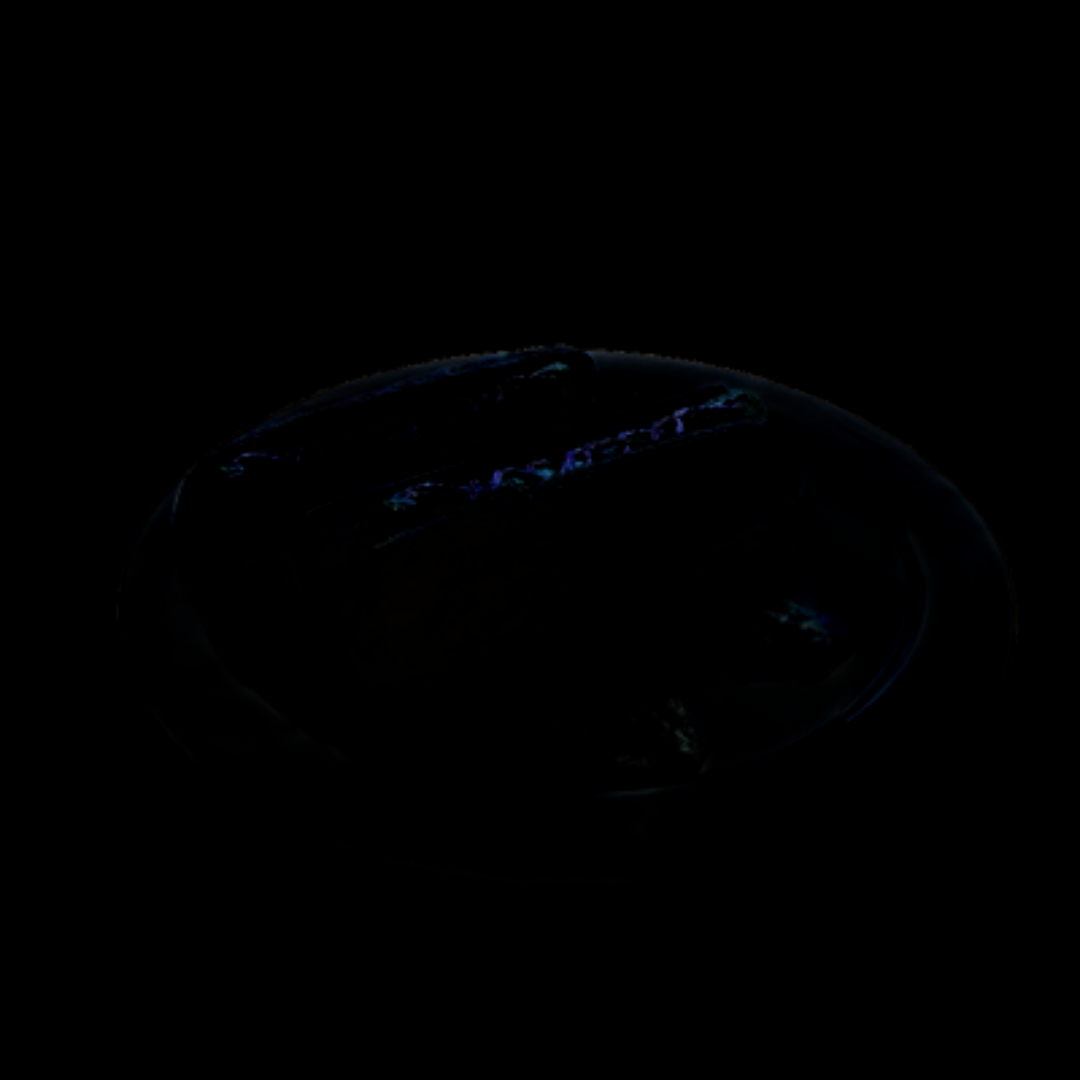}}
\subfloat{\includegraphics[width=0.24\linewidth]{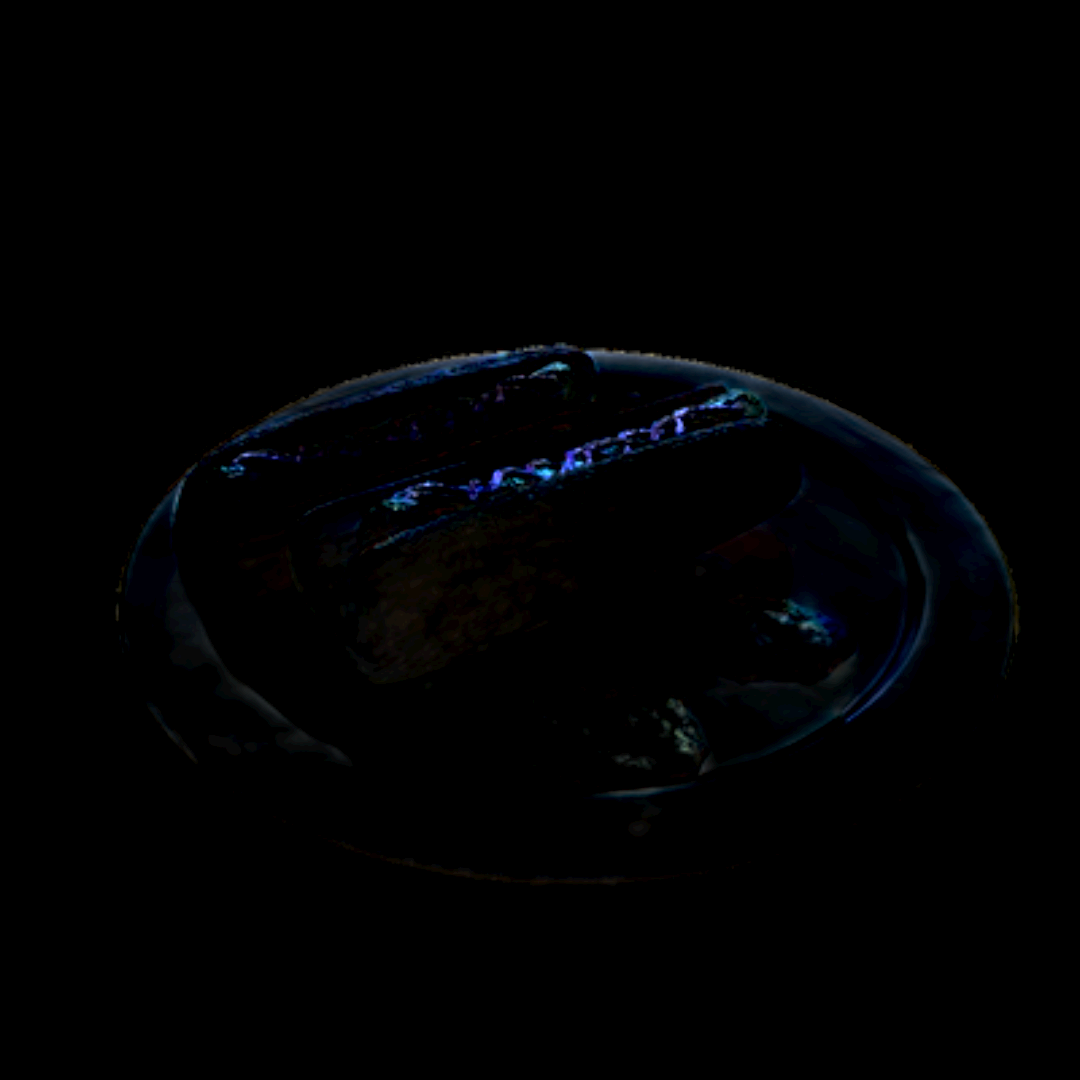}}

\subfloat{\includegraphics[width=0.24\linewidth]{figures/drums_diffuse.png}}
\subfloat{\includegraphics[width=0.24\linewidth]{figures/drums_specular.png}}
\subfloat{\includegraphics[width=0.24\linewidth]{figures/drums_specular_only.png}}
\subfloat{\includegraphics[width=0.24\linewidth]{figures/drums_specular_only_bright.png}}

\subfloat{\includegraphics[width=0.24\linewidth]{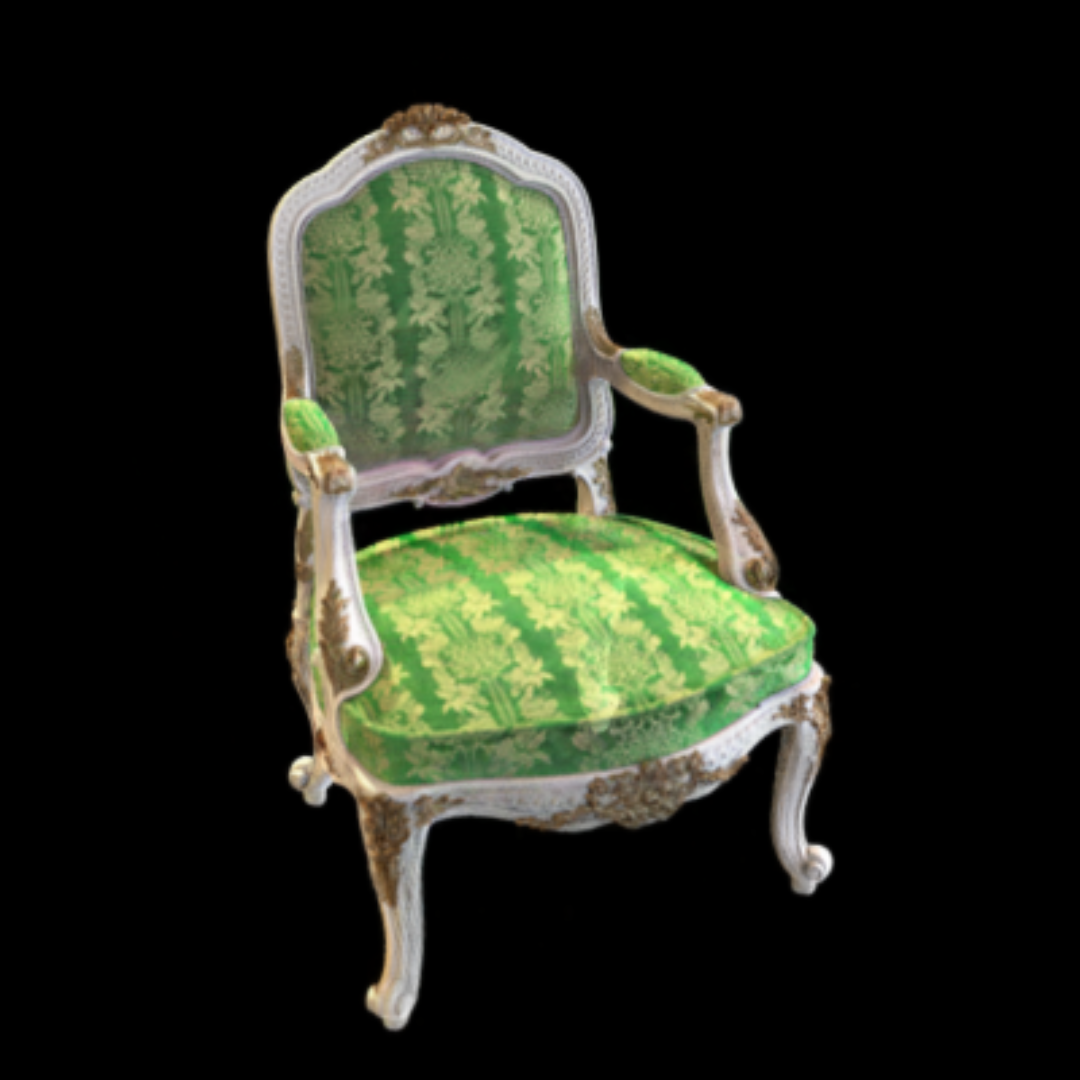}}
\subfloat{\includegraphics[width=0.24\linewidth]{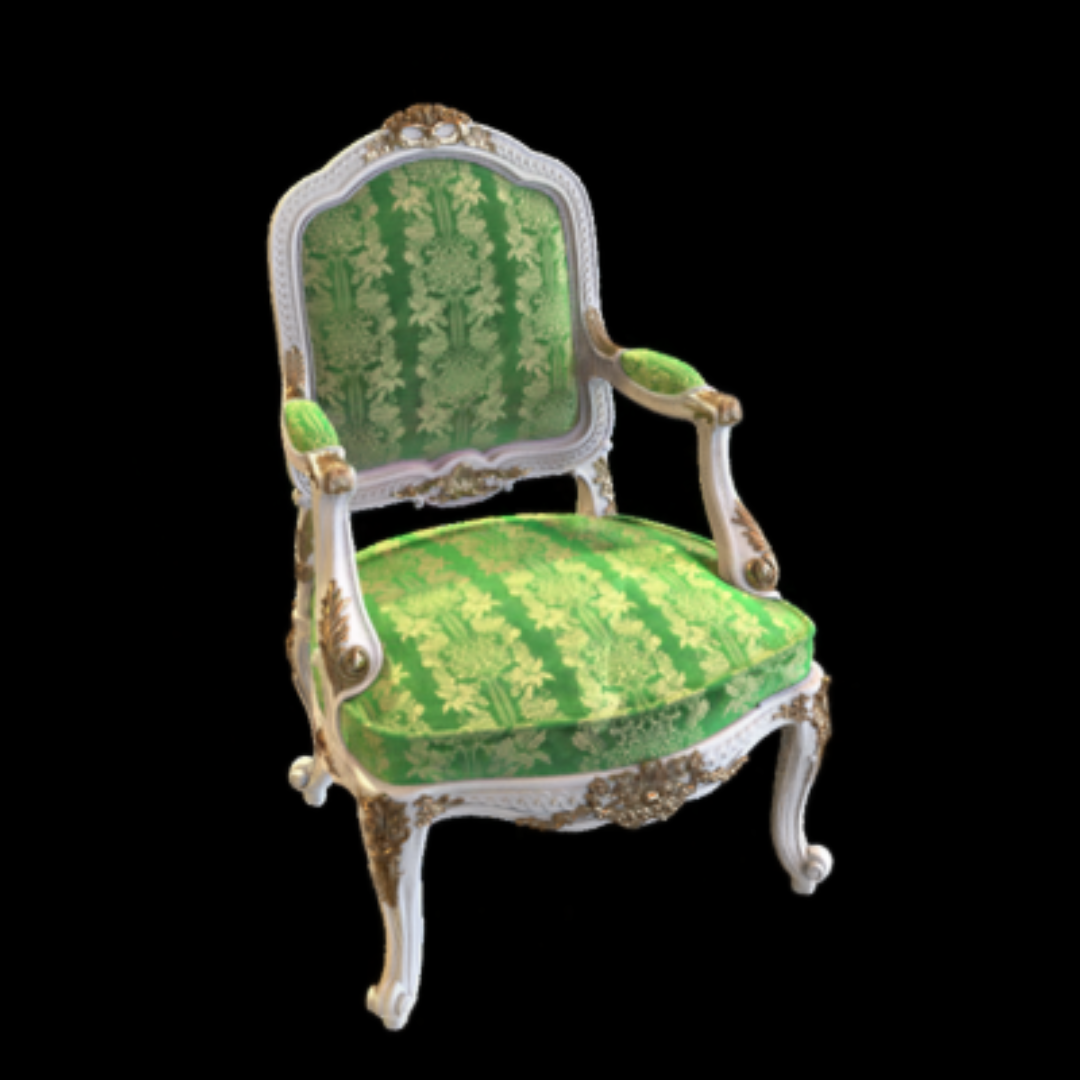}}
\subfloat{\includegraphics[width=0.24\linewidth]{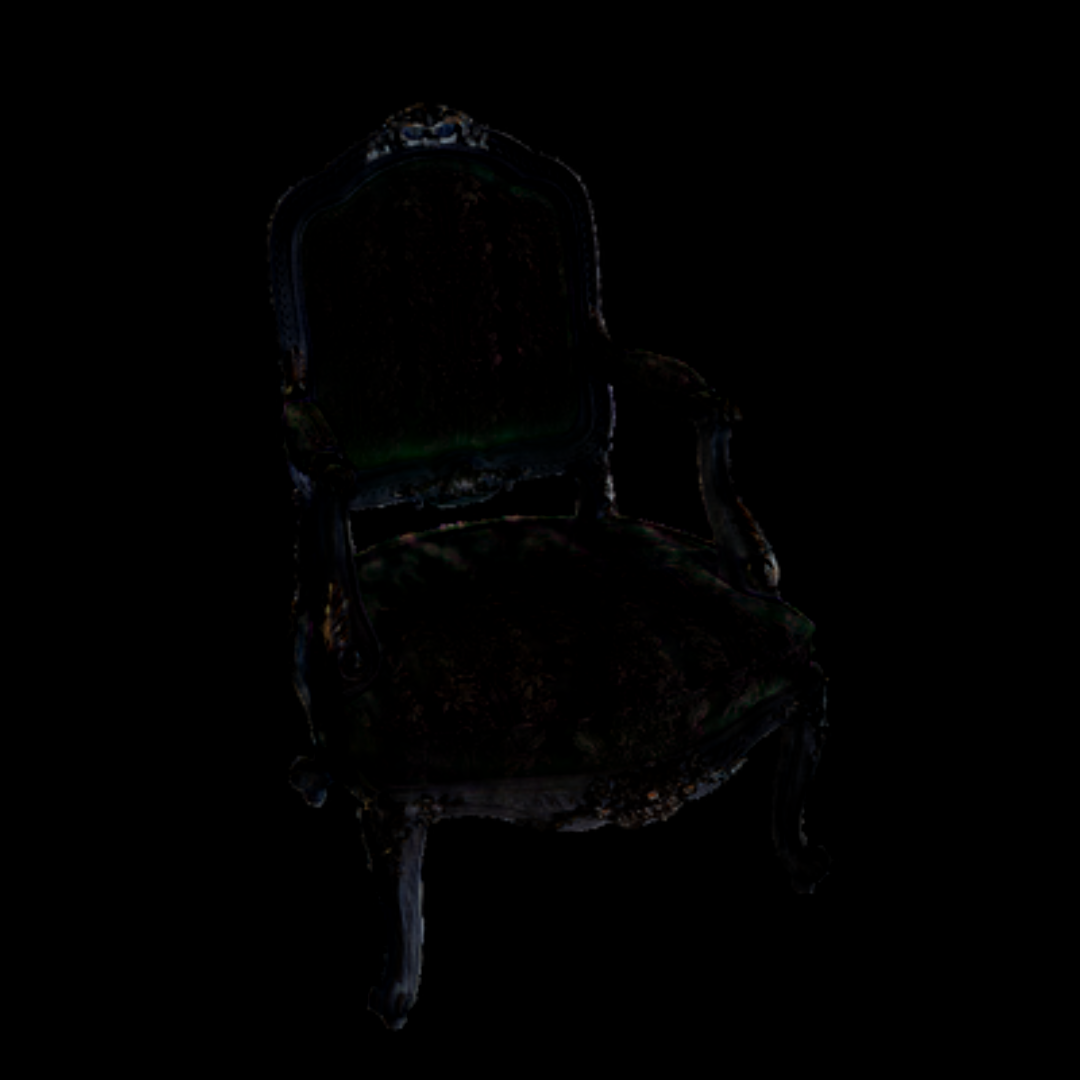}}
\subfloat{\includegraphics[width=0.24\linewidth]{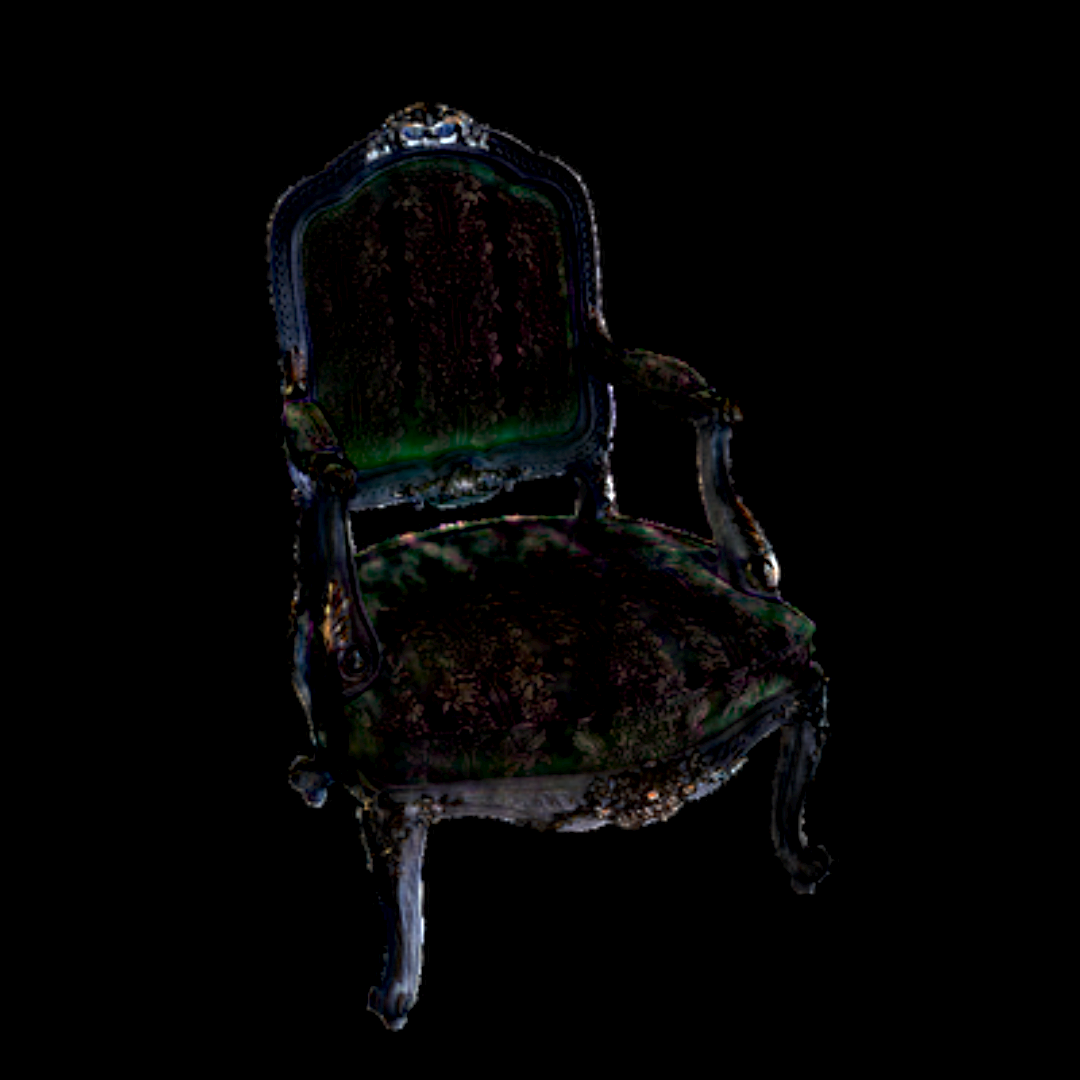}}

\subfloat{\includegraphics[width=0.24\linewidth]{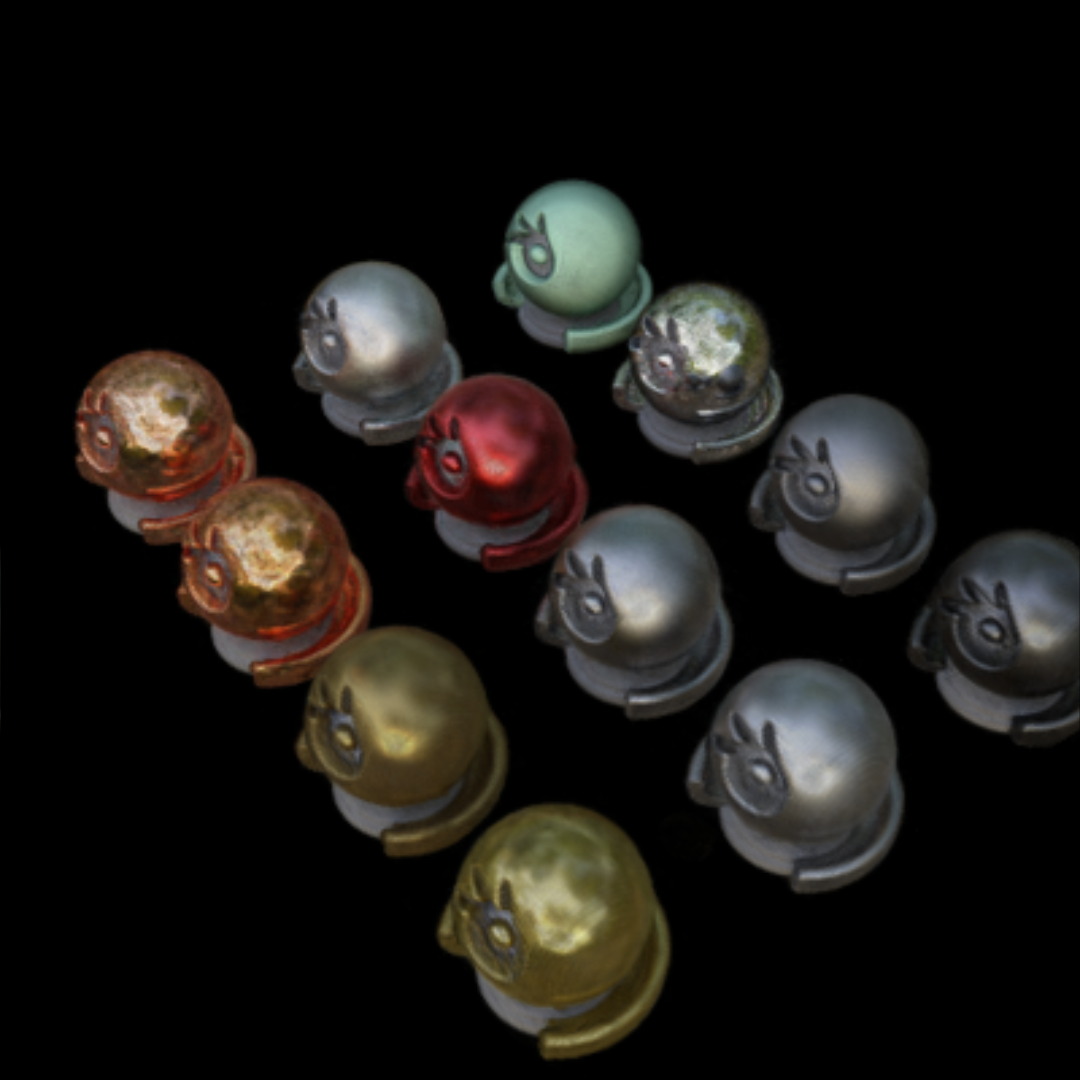}}
\subfloat{\includegraphics[width=0.24\linewidth]{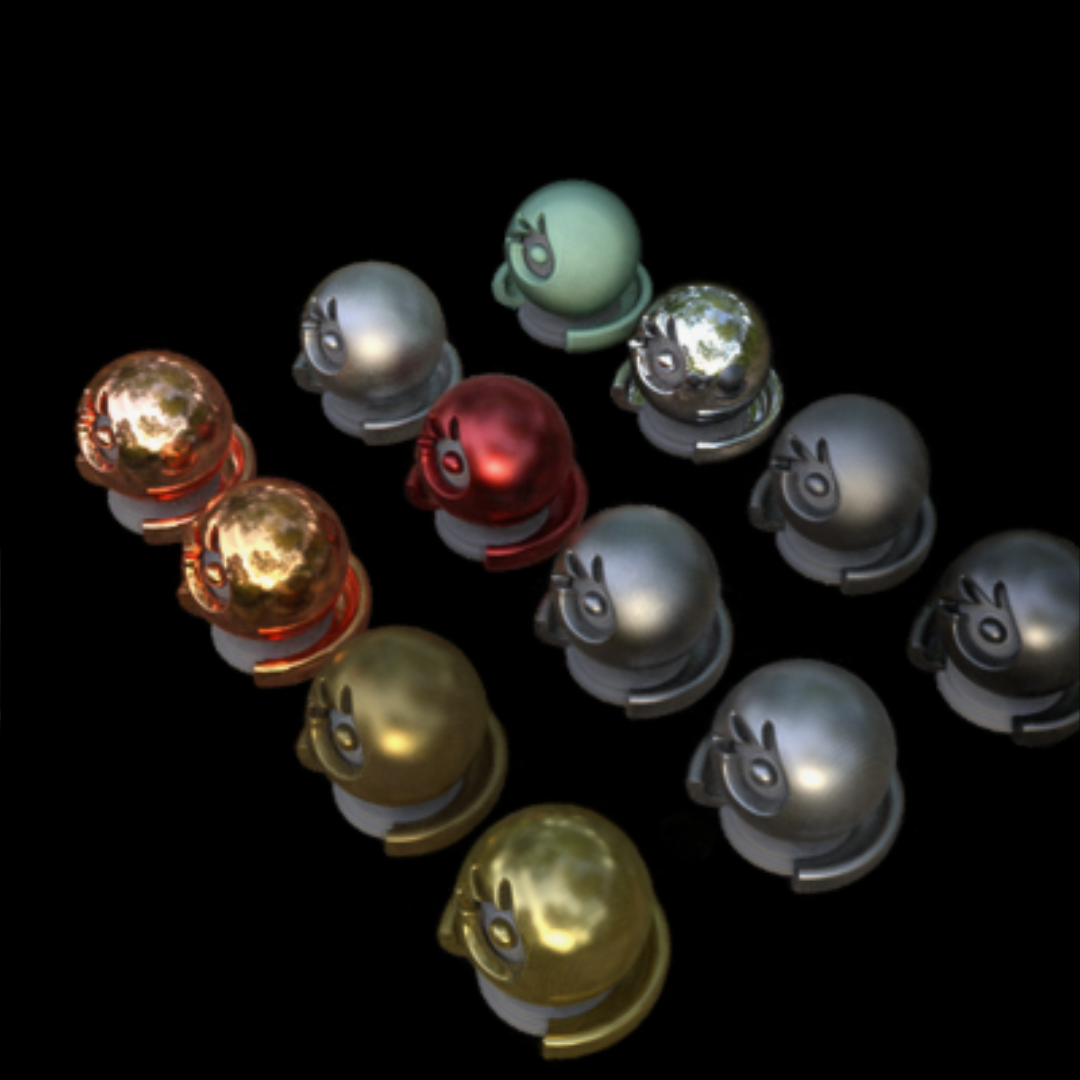}}
\subfloat{\includegraphics[width=0.24\linewidth]{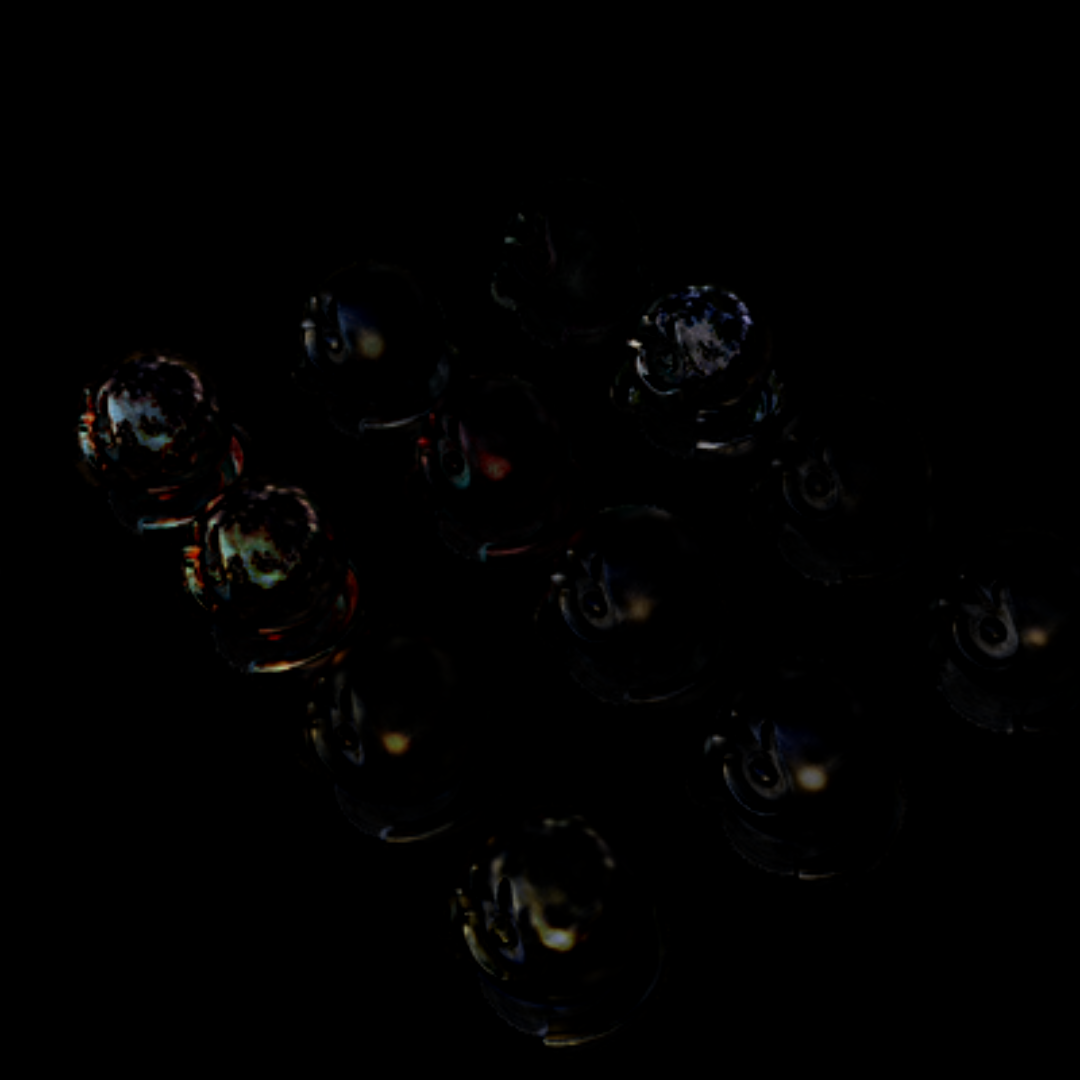}}
\subfloat{\includegraphics[width=0.24\linewidth]{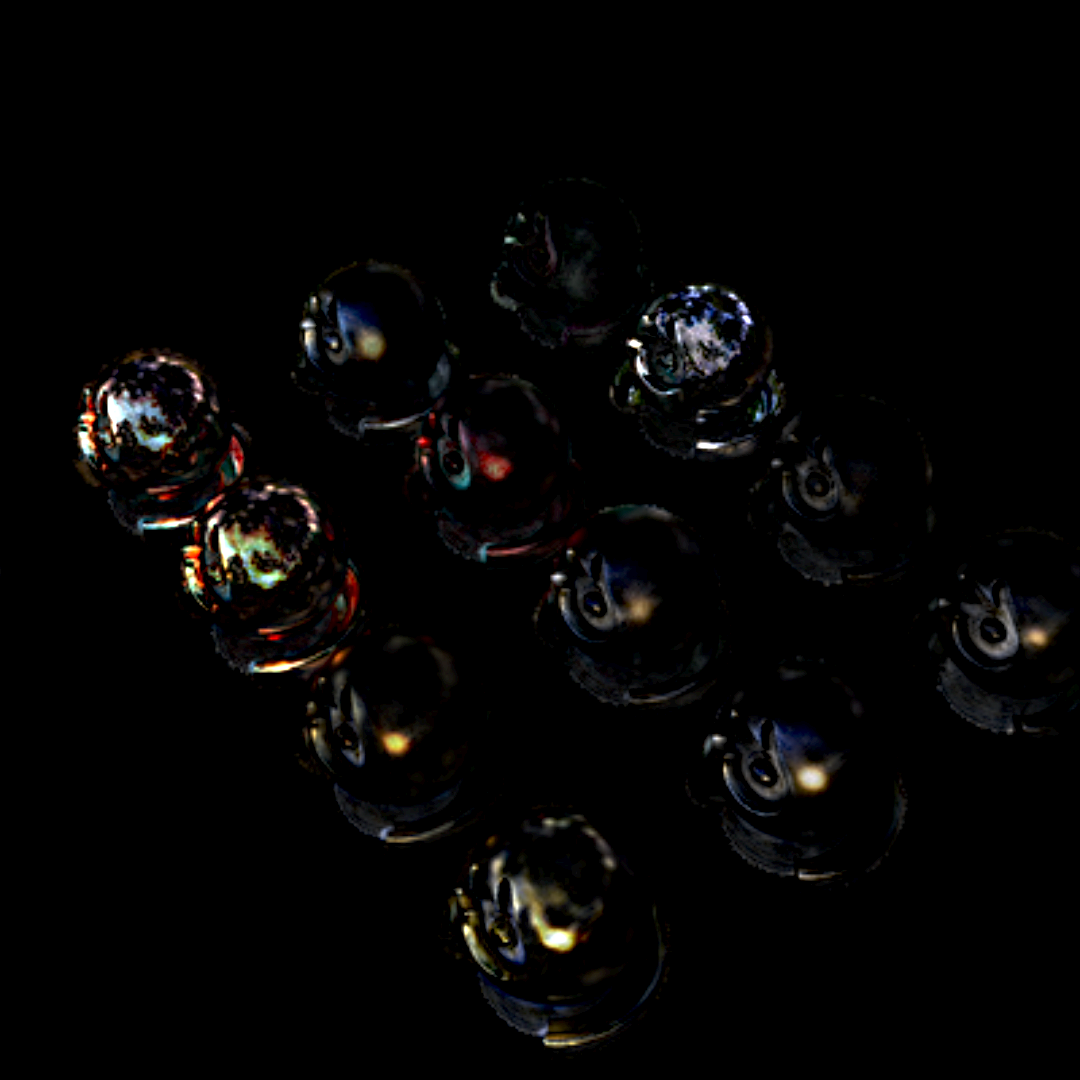}}

\subfloat{\includegraphics[width=0.24\linewidth]{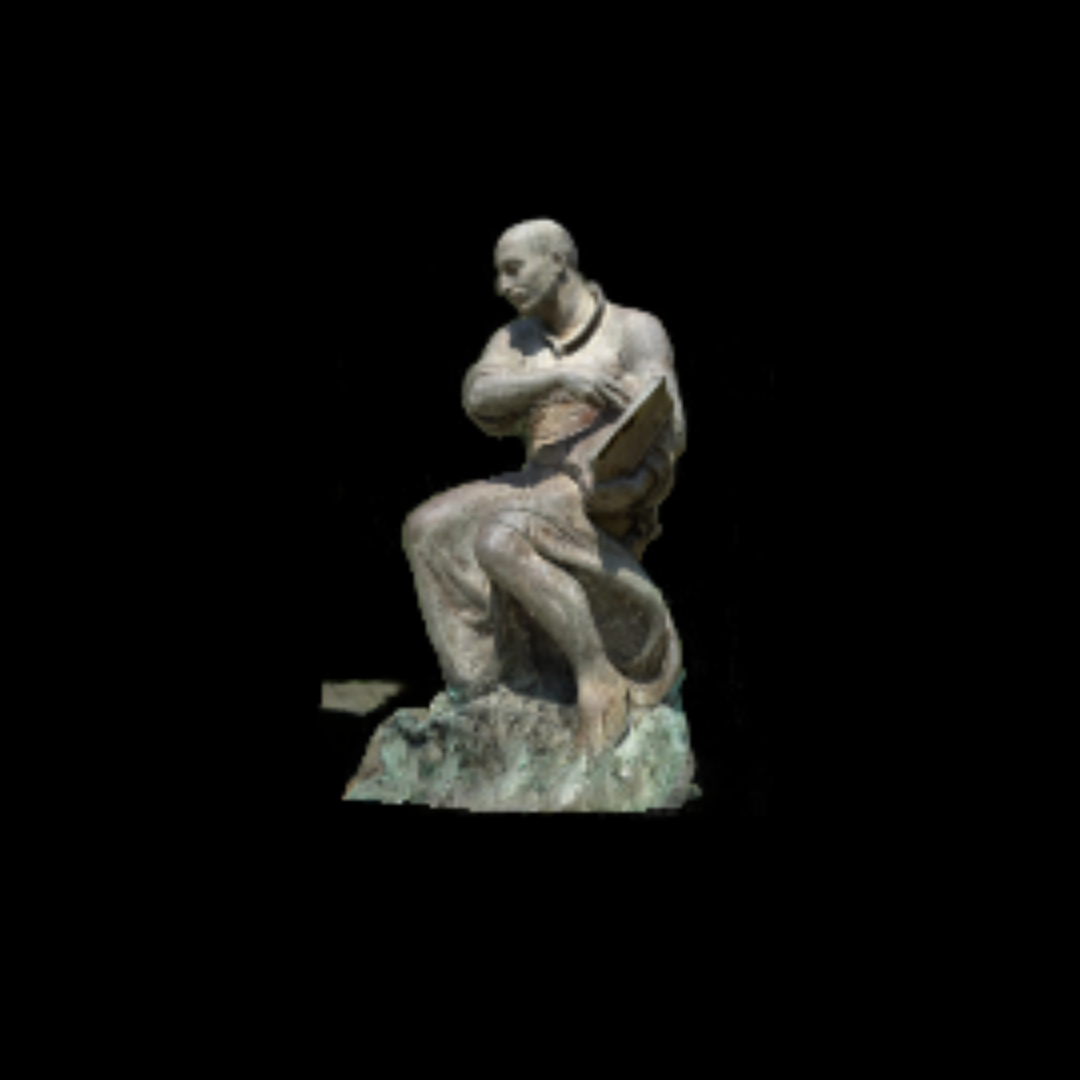}}
\subfloat{\includegraphics[width=0.24\linewidth]{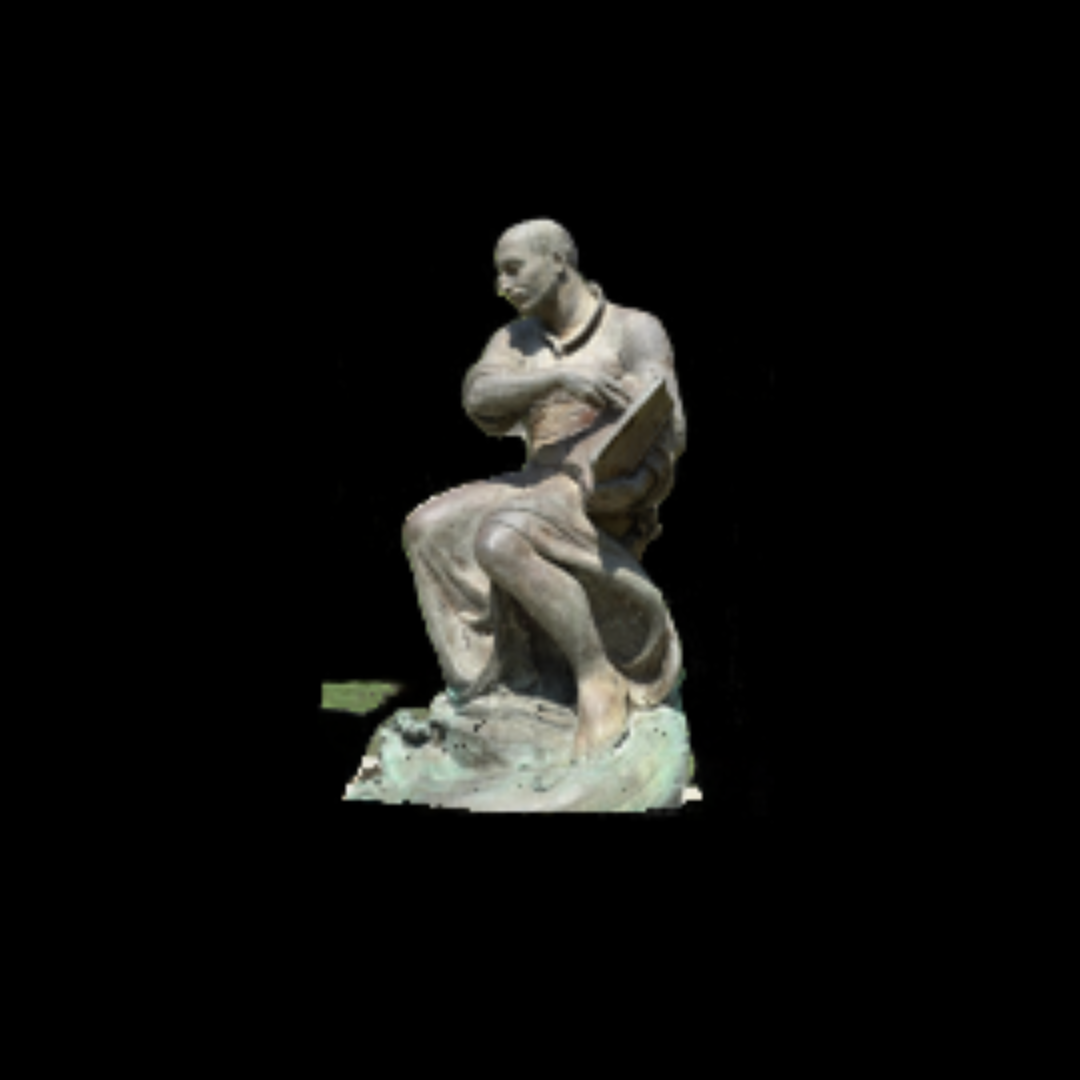}}
\subfloat{\includegraphics[width=0.24\linewidth]{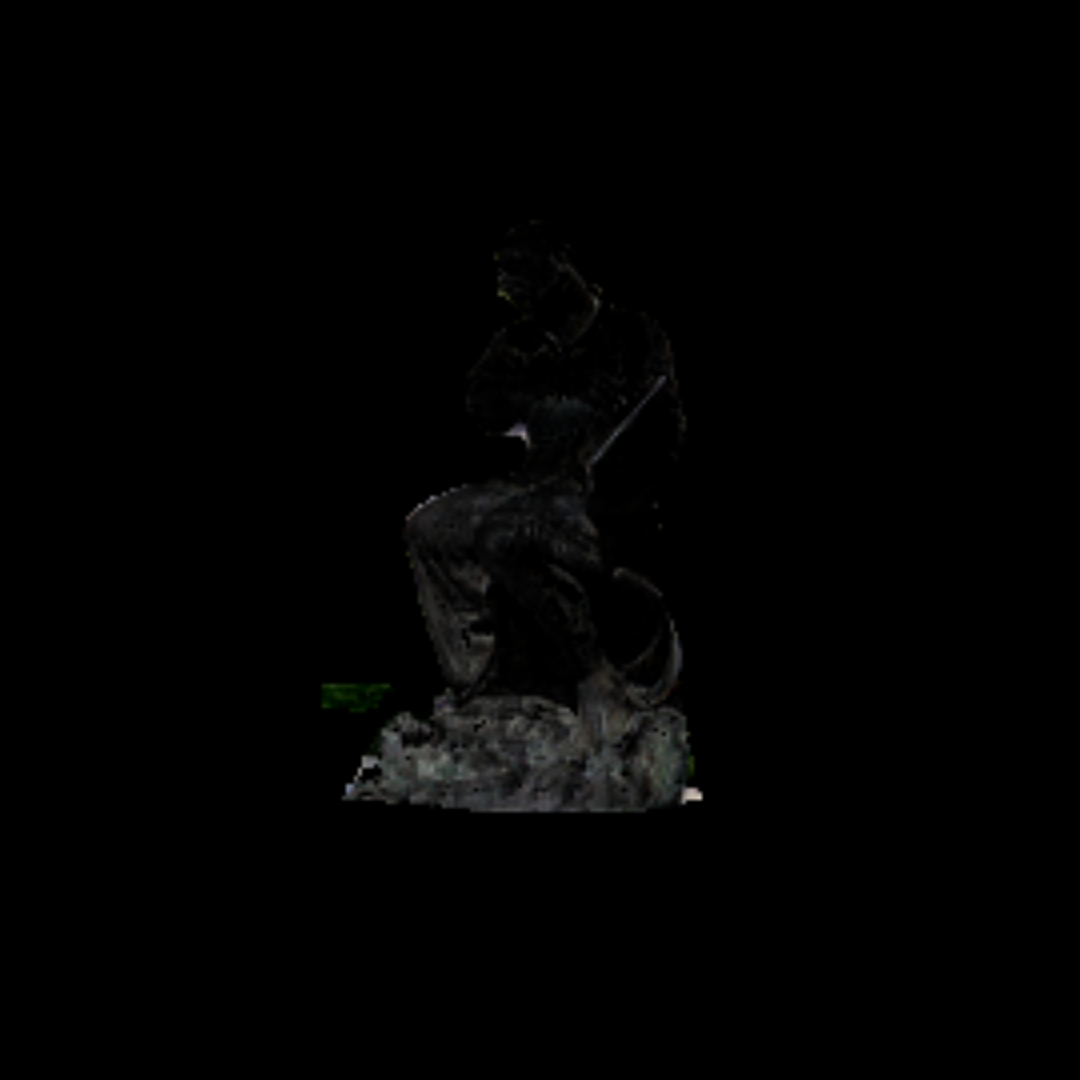}}
\subfloat{\includegraphics[width=0.24\linewidth]{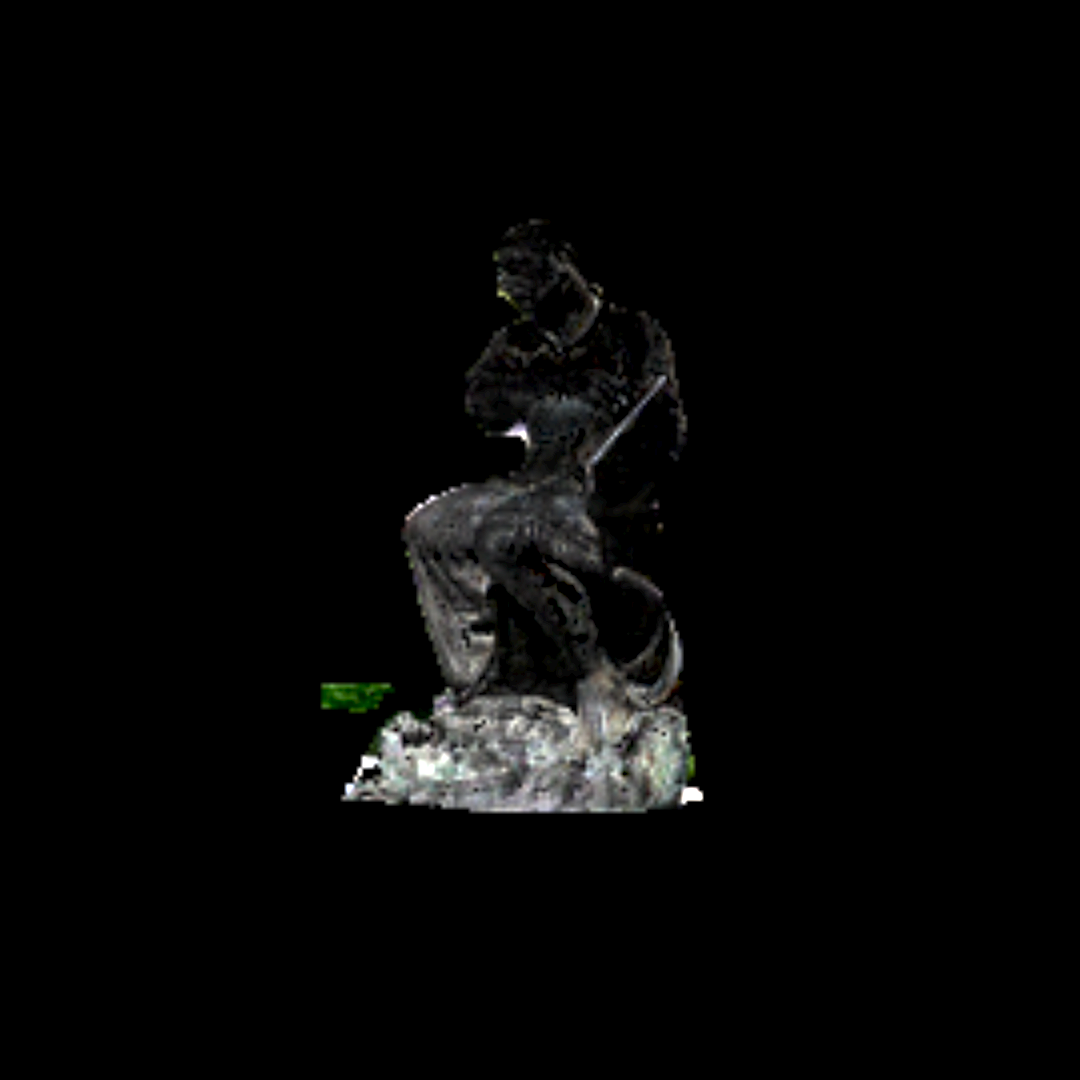}}

\subfloat{\includegraphics[width=0.24\linewidth]{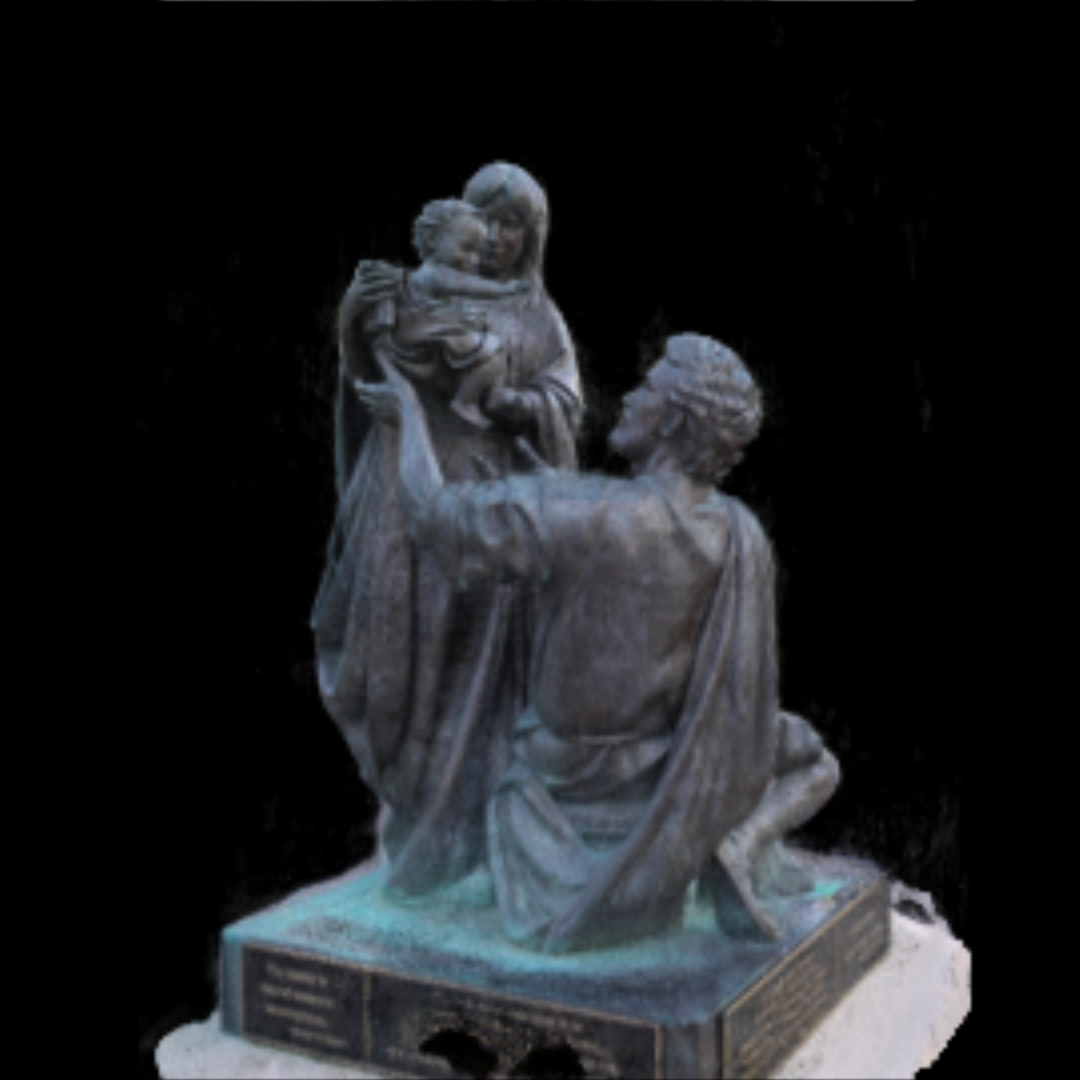}}
\subfloat{\includegraphics[width=0.24\linewidth]{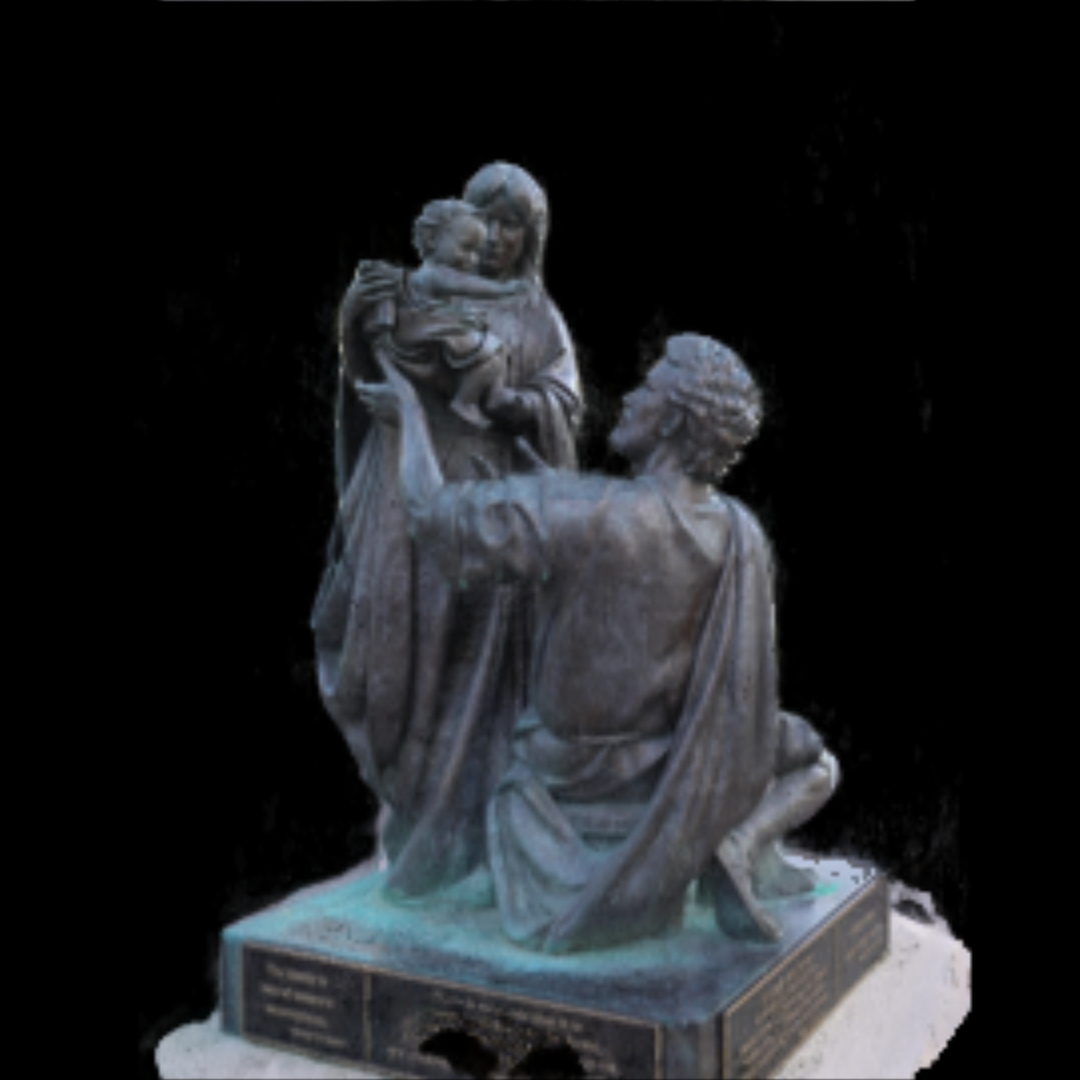}}
\subfloat{\includegraphics[width=0.24\linewidth]{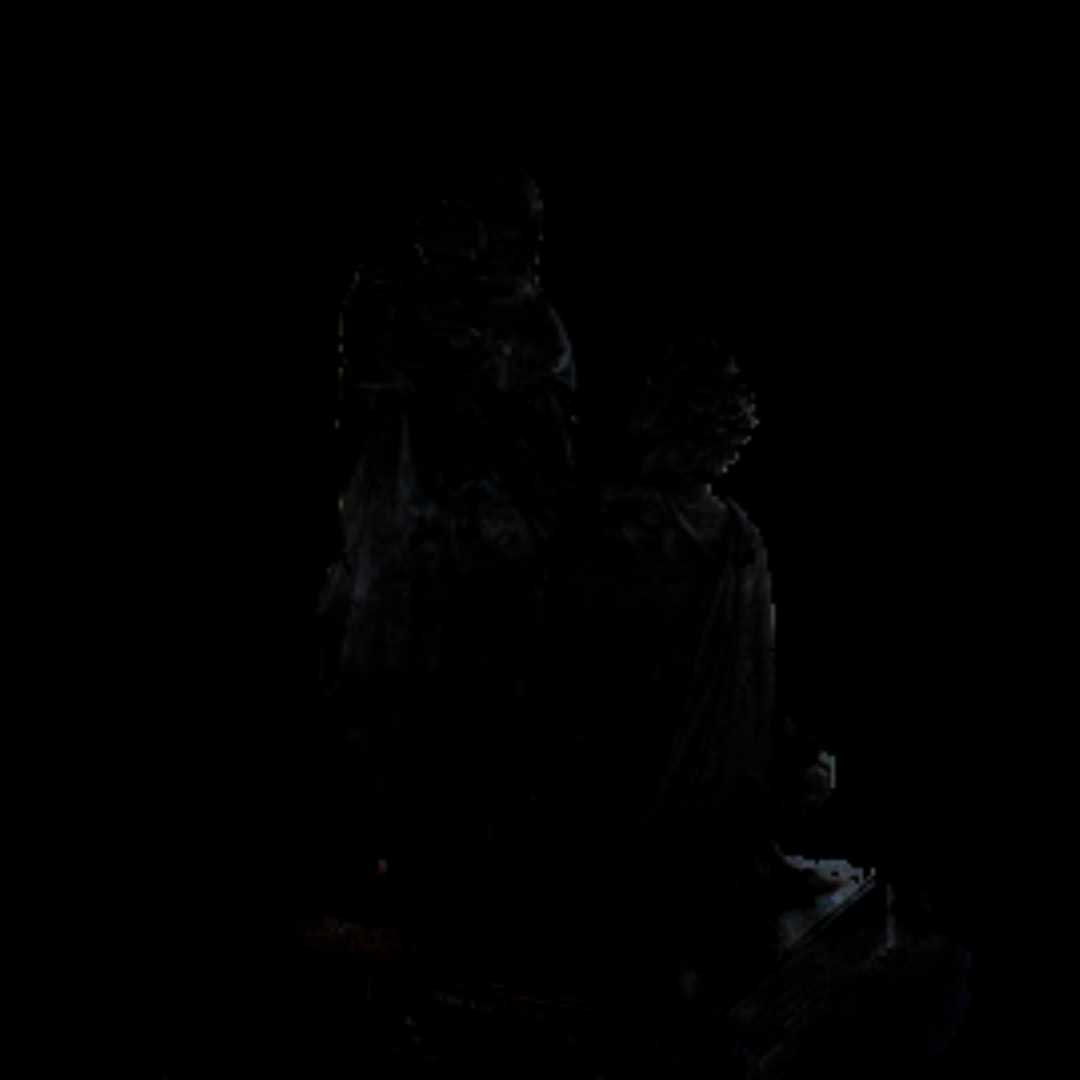}}
\subfloat{\includegraphics[width=0.24\linewidth]{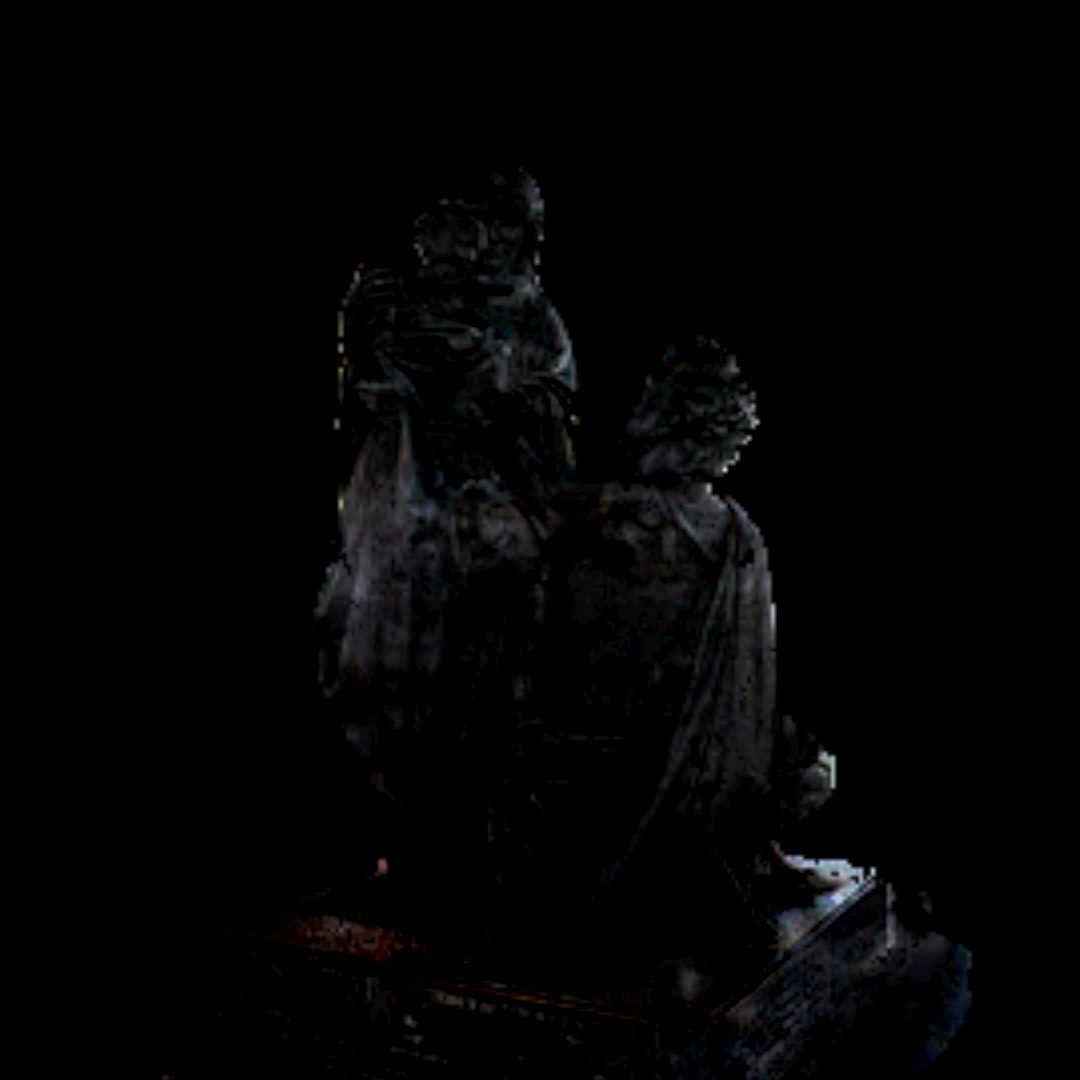}}

\label{fig:results}
\caption{Results from our algorithm. From left to right: Lambertian color (from volume), Lambertian + view-dependent color, 
view-dependent color only,  view-dependent color brightened for easier inspection. }
\end{figure}

\noindent In Figure~\ref{fig:results}, we test our solution on a number of different 360\textdegree\ scenes from the artificial NERF data set, and from the 
Tanks and Temples dataset of captured scenes\cite{Mildenhall2020}\cite{Knapitsch2017}. We use four hierarchical levels with $256^3$ for 
the hightest level. In most cases, it is possible to separate out view-dependent information, as can be seen in the rendering of the view-dependent parts on the right.

\begin{table}[]
\centering
\begin{tabular}{@{}|l|ll|ll|@{}}
\toprule
      \multicolumn{5}{|c|}{Timings (seconds)}  \\ \midrule
         & \multicolumn{2}{c|}{w/ planes} & \multicolumn{2}{c|}{w/o planes} \\ \midrule
lego     & \multicolumn{2}{c|}{154}       & \multicolumn{2}{c|}{194}        \\
ficus    & \multicolumn{2}{c|}{107}       & \multicolumn{2}{c|}{208}        \\
hotdog   & \multicolumn{2}{c|}{267}       & \multicolumn{2}{c|}{355}        \\
drums    & \multicolumn{2}{c|}{121}       & \multicolumn{2}{c|}{273}        \\
chair    & \multicolumn{2}{c|}{90}        & \multicolumn{2}{c|}{229}        \\
material & \multicolumn{2}{c|}{243}       & \multicolumn{2}{c|}{478}        \\
ignatius & \multicolumn{2}{c|}{228}       & \multicolumn{2}{c|}{172}        \\
family   & \multicolumn{2}{c|}{154}       & \multicolumn{2}{c|}{131}        \\ \bottomrule
\end{tabular}
\vspace*{3mm}
\caption{ \NoCaseChange{Timings when reconstructing with our two additions (left column) and without (right column). For the artificial scenes
our addition becomes a net speed up, even though additional computation is performed. In the real scenes (bottom two) the
extra computation time is more noticeable.} }
\label{tab:timings}
\end{table}

The scenes are reconstructed using a Nvidia GTX 1080 graphics card. In Table~\ref{tab:timings}, we see reconstruction times for when using
the additions described in this paper compared to baseline PERF. To make a fair comparison, Equation~\ref{eq:cauchy} is added with a 
constant $w = 0.0001$ to the baseline case.

For the artificial scenes, we actually have a higher performance due to faster convergence of the reconstruction. For the two real scenes, which have more camera poses than the artificial scenes, the extra overhead is about 20-30\%.

\section{Discussion}
\noindent We present a novel way of mitigating the shape-radiance ambiguity in volume reconstruction with view-dependent colors.
Our method requires little extra memory and has a small overall impact on performance, 
and might even be beneficial for some scenes due to improved convergence.
The resulting reconstruction can be used as is with geometry and Lambertian colors, or be further augmented with view-dependent 
information at a later  stage. Given a correct start geometry, a high-resolution function describing view-dependence can be used, without problems with the shape-radiance ambiguity.

\subsection{Limitations and Future Work}
\noindent Our proposed addition is able to successfully capture deviating radiance compared to the Lambertian volume, such as a highlight
seen from one or a few cameras. In more complex cases,  our small view-dependent function can help with convergence to the correct 
surface. 
The next step, which we have not studied in this paper, is to use this separation of Lambertian and view-dependent colors
to either explicitly reconstruct the geometry with e.g. marching cubes, and/or to fit a high resolution view-dependent function
to describe the incoming radiance with high accuracy. 

%However, as shown in Figure~XX, it can be difficult for the algorithm to properly separate the diffuse and specular components in certain 
%difficult scenarios. 

% use section* for acknowledgment
\ifCLASSOPTIONcompsoc
  % The Computer Society usually uses the plural form
  \section*{Acknowledgments}
\else
  % regular IEEE prefers the singular form
  \section*{Acknowledgment}
\fi
\noindent This work was supported by the Swedish Research Council under
Grant 2014-4559.

\bibliographystyle{IEEEtran}
\bibliography{refs}

\end{document}